\documentclass{article}

\PassOptionsToPackage{numbers, compress}{natbib}
\usepackage[main, final]{neurips_2026}

\usepackage[table]{xcolor}

\usepackage[utf8]{inputenc} % allow utf-8 input
\usepackage[T1]{fontenc}    % use 8-bit T1 fonts
\usepackage{hyperref}       % hyperlinks
\usepackage{url}            % simple URL typesetting
\usepackage{booktabs}       % professional-quality tables
\usepackage{amsfonts}       % blackboard math symbols
\usepackage{nicefrac}       % compact symbols for 1/2, etc.
\usepackage{microtype}      % microtypography
\usepackage{natbib}
\usepackage{graphicx}
\usepackage{adjustbox}
\usepackage{amsmath}
\usepackage{wrapfig}
\usepackage{longtable}
\usepackage{comment}
\usepackage{xspace}
\usepackage[textsize=tiny]{todonotes}
\usepackage[most]{tcolorbox}
\usepackage{makecell}
\usepackage{collcell}
\usepackage{booktabs}
\usepackage{multirow}
\usepackage{tcolorbox}
\usepackage{enumitem}
\usepackage{bbding}
\usepackage{subcaption}
\usepackage{caption}
\usepackage{xcolor}
\usepackage{colortbl}

\tcbuselibrary{breakable, skins}
\newtcolorbox{promptbox}[1][]{
  breakable,
  enhanced,
  colback=gray!5,
  colframe=gray!40,
  fonttitle=\bfseries\small,
  title={#1},
  left=4pt, right=4pt, top=4pt, bottom=4pt,
  before upper={\setlength{\parskip}{6pt}\setlength{\parindent}{0pt}}
}

\newcommand{\ie}{\textit{i.e.}}
\newcommand{\eg}{\textit{e.g.}}
\newcommand{\shor}{\raisebox{-0.75pt}{\includegraphics[height=0.85em]{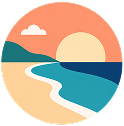}}}
\usepackage{amsmath,amsfonts,bm}

\def\eqref#1{equation~\ref{#1}}
\def\1{\bm{1}}

\DeclareMathAlphabet{\mathsfit}{\encodingdefault}{\sfdefault}{m}{sl}
\SetMathAlphabet{\mathsfit}{bold}{\encodingdefault}{\sfdefault}{bx}{n}

\definecolor{lightblue}{RGB}{224,236,247}
\definecolor{deepblue}{RGB}{9,46,107}

\title{Towards Direct Evaluation of Harness Optimizers\\via Priority Ranking}

\author{%
    Kai Tzu-iunn Ong$^{1}\thanks{KT Ong, M Kang, and D Choi are the co-first authors.}$~
    ~ Minseok Kang$^{1*}$
    ~ Dongwook Choi$^{1*}$
    ~ Junhee Cho$^{1}$\\
    \textbf{Seungju Kim}$^{1}$
    \quad \textbf{Seungwon Lim}$^{1}$
    \quad \textbf{Geunha Jang}$^{1}$
    \quad \textbf{Minwoo Oh}$^{1}$
    \quad \textbf{Bogyung Jeong}$^{1}$\\
    \quad \textbf{Sunghwan Kim}$^{1}$
    \quad \textbf{Taeyoon Kwon}$^{1}$
    \quad \textbf{Jihye Han}$^{2}$
    \quad \textbf{Juyoung Wy}$^{2}$
    \quad \textbf{Jinyoung Yeo}$^{1}$ \\\\
  $^{1}$LangAGI Lab, Department of Artificial Intelligence, Yonsei University\\
  $^{2}$Samsung Electronics
}

\begin{document}

\maketitle

\begin{abstract}
    \textit{Harness optimization} enables automated agent creation by having an optimizer agent iteratively update the harness of target agents. Despite its success, current studies evaluate optimizers solely by observing target agents' performance gains. 
  This indirect end-improvement evaluation neglects optimizers' actions at intermediate steps, which are often erroneous and hinder agent performance in deployment. Therefore, it is unclear whether harness optimization is driven by optimizers' informed update actions or simply trial-and-error. 
  This necessitates direct evaluation of harness optimizers.
  However, evaluating harness optimizers directly is non-trivial and costly due to the lack of oracle harnesses. To address this, we present a simple, low-cost design to directly evaluate them, namely \textit{priority ranking}.
  By asking harness optimizers to rank components (\eg, tools) in a given harness by their potential to improve/hinder agent performance when updated, our design quantifies optimizer ability at the step level \textit{without expensive rollouts or manual examination}. More importantly, optimizers' ranking performance correlates with their ability to improve agents in actual multi-step harness optimization, establishing priority ranking as a reliable indicator of optimization ability. Priority ranking is enabled by \shor~\textsc{Shor}, a collection of 182 human-verified optimization scenarios spanning across domains, designs, and time stages. \href{https://github.com/k59118/Harness_Optimizer_Evaluation}{\textbf{[\texttt{CODE\,\&\,DATA}]}}

\end{abstract}

\section{Introduction}
Large language model (LLM) agents have been assisting humans across various domains, such as software engineering (SWE) and customer service~\citep{yao2022react, chae2023dialogue, kwon2024large, chae-etal-2024-coffee, hu-etal-2025-compileagent, ong-etal-2025-towards, tian2026swe, kwon2025embodied, lee2026self, kim2026persona2web}. Their performance largely depends on the design of the surrounding ``\textit{\textbf{harness}}''~\textemdash~the software layer around the LLM brain that manages its workflow, context, and external interactions (\eg, tools)~\citep{Anthropic_2026}. 
For instance, \citet{chae2025web} shows that changing how observations are perceived improves web agents' success rate (SR) by 10\%p, and workflow changes create up to 6$\times$ performance gap in SWE agents using the same LLM brain~\citep{tian2026swe}.

Despite the success of agents, manually crafting harnesses is costly and hard to scale.
Thus, recent studies have pivoted towards automating \textit{harness optimization}~\citep{hao2026recreate, hu2024automated, zhang2024aflow, agrawal2025gepa, suzgun-etal-2026-dynamic, zhang2025agentic, ursekar2026vero, zhang2026hyperagents, lee2026meta, yuksekgonul2025optimizing}.
Generally, it is achieved by having an agent act as the optimizer (\ie, outer loop), iteratively updating the harness of a target agent (\ie, inner loop) based on its task performance and trajectories~\citep{hao2026recreate}. 
By converting massive execution logs into actionable insights and fixing task-specific bottlenecks step-by-step, harness optimization has enabled better agent performance than hand-written harnesses \citep{zhang2025agentic, ursekar2026vero, zhang2026hyperagents, lee2026meta, yuksekgonul2025optimizing}. 

Within this trend, existing studies treat harness optimizers as black boxes~\textemdash~the effectiveness of optimizers is judged solely by observing target agents' improvement after the entire optimization process. This simplification, while straightforward, neglects optimizer actions during the optimization process. As previously reported~\citep{hao2026recreate, ursekar2026vero, zhang2026hyperagents}, harness optimizers often make improper harness updates at intermediate steps (\eg, building a workflow that causes the agent to unconditionally end tasks when format errors occur), hindering final agent performance. Yet, common agent-centric, end-improvement evaluation provides no signals about these optimizer actions, leaving it unclear whether harness optimization is driven by informed optimizer actions or trial-and-error.

\begin{figure}[t]
    \centering
\includegraphics[width=1\linewidth]{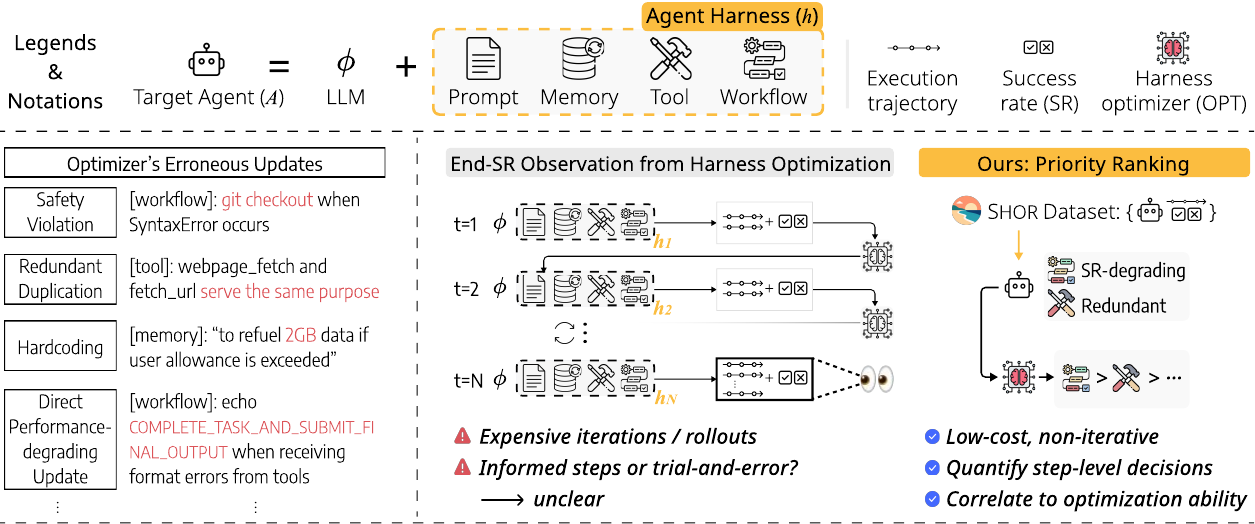}
    \caption{(Right) End-improvement observation vs. priority ranking. Our design quantifies optimizers' ability cost-and time-effectively and directly, whereas existing evaluations require running the entire optimization process and offer limited insights; (Left) Examples of erroneous harness updates.}
    \label{fig:motivation_large}
    \vspace{-17.5pt}
\end{figure}

At this point, we argue that evaluating harness optimizers should go beyond agent performance, and include a direct focus on optimizer ability to prioritize effective update actions over detrimental ones at the step level.
However, directly assessing optimizers' update actions within the actual optimization process is non-trivial.
Realistically, at any time step, there is \textbf{no oracle next harness}.
Judging whether an optimizer has prioritized an effective update requires either human assessments, or massive exploration of updatable harness components (and how they affect agents). Doing so at scale further incurs \textbf{extensive rollouts} across tasks, which are time- and cost-consuming as harness optimization involves coupled execution of the optimizer and agent.

This work is a first step to address these bottlenecks. We propose a simple evaluation design that quantifies optimizer ability time- and cost-efficiently, namely \textit{priority ranking} (Figure~\ref{fig:motivation_large}, right). Crucially, we show that priority ranking serves as a reliable predictor of optimizers' actual optimization ability.
The core of this task is to have an optimizer rank the components (\ie, \textit{prompt}, \textit{memory}, \textit{workflow}, and \textit{tool}) of a given harness based on their potential to improve/harm agent performance when updated. 
For example, if changing the system prompt is expected to bring greater improvement than adding a new tool, \texttt{prompt} should be ranked higher than \texttt{tool}.
This design sidesteps both bottlenecks in directly evaluating optimizers:
(i) ranking requires only \textbf{relative priorities} among components, rather than absolute next harnesses or human assessment;
(ii) ranking is a \textbf{non-iterative text generation task}, eliminating costly, time-consuming rollouts; 
To enable priority ranking, we curate \shor~\textsc{Shor}, a collection of 182 human-verified optimization scenarios across domains and time stages. Ranking labels are annotated via a collaborative effort of SOTA agents and quality control criteria.
Our contributions can be summarized as follows:
\begin{itemize}[leftmargin=*, topsep=2pt,itemsep=3pt,parsep=0pt]
    \item Within the trend of harness optimization, we go beyond agent performance and \textbf{take a first step} towards the direct evaluation of harness optimizers.
    \item \textbf{Performance in priority ranking correlates with optimizers' actual ability} to improve agents via multi-step harness optimization ($\rho$ = 0.602 in-domain; 0.632 out-of-domain), establishing priority ranking as a reliable pre-deployment proxy for actual optimization performance.
    \item Priority ranking via \textsc{Shor} is at least \textbf{8$\times$ cheaper} and \textbf{17$\times$ faster} than the common practice, providing a practical and efficient alternative for comparing optimizers.
    \item Further analysis shows that explicitly addressing optimizers' awareness of optimization priorities largely improves their performance in correcting flawed harnesses. Priority ranking thus serves as more than an evaluation design but also \textbf{an actionable insight for developing harness optimizers}.
    
\end{itemize}

\section{Related Work}
\label{related_work}

\paragraph{Self-evolving LLMs.}
While harness optimization is relatively new, efforts to automatically improve LLM systems have been made for years~\citep{yuksekgonul2025optimizing, yang2023large, ong2024large, novikov2025alphaevolve, wang2025thetaevolve, yue2026dr}:
Early work like OPRO~\citep{yang2023large} has the LLM to come up with new solutions based on its previous solutions and their scores; \citet{shinn2023reflexion} model a reflection process on past failures to improve next prediction; TextGrad \citep{yuksekgonul2025optimizing} treats LLMs as hidden layers, eliciting natural language (NL) gradients for prompt optimization. 
These exhibit how LLMs can self-evolve through experiences. Upon it, Expel \citep{zhao2024expel} applies an insight extraction module and memory retrieval to better elicit strategic cues from experience. Dynamic Cheatsheet~\citep{suzgun-etal-2026-dynamic} also adapts such structural memory management, yielding near-perfect Acc in arithmetic tasks. Also, in the dialogue domain, \citet{kim-etal-2025-principles} show that LLMs can derive appropriate conversational principles for emotional support by analyzing successful and failed past attempts.

\paragraph{Harness optimization.}
With the rise of agents, self-evolution has scaled to the entire harness surrounding the LLM brain, known as \textit{harness optimization}~\citep{hu2024automated, zhang2024aflow, agrawal2025gepa, qu2026coral}. 
Generally, this line of work uses an LLM or agent as an optimizer to iteratively update the target agent's harness.
GEPA~\citep{agrawal2025gepa} uses an optimizer to refine or merge agents across tasks based on their trajectories and previous performance. ACE~\citep{zhang2025agentic} uses a Dynamic Cheatsheet-inspired optimizer for test-time memory optimization to prevent context collapse. ADAS~\citep{hu2024automated} treats harness optimization as a programming task and updates \texttt{forward()}s in the code, optimizing the order or conditional branches of predefined modules; AFlow \citep{zhang2024aflow} further includes MCTS~\citep{chaslot2010monte} to find the best workflow in defined candidates. 
This year, ReCreate~\citep{hao2026recreate} takes a step further, addressing harness creation from scratch via dual-level (task and domain) optimization with lower costs than methods built on predefined harnesses.
Another stream of work focuses on policy training~\citep{wang2025scoreflow, xu2025robustflow, gao2025flowreasoner, kong2026workflow, yue2026static}: \citet{wang2025scoreflow} propose a score-based direct preference optimization (DPO)~\citep{rafailov2023direct} and train a harness generator; Workflow-R1~\citep{kong2026workflow} frame workflow construction as multi-turn, NL-based decision-making and uses a group subsequence policy optimization (GSsPO) to train an optimizer. 
While these often rely on NL optimization signals, \citet{ursekar2026vero} and \citet{lee2026meta} formulate harness optimization as open-ended coding tasks, adopting coding agents to optimize the harness. 
Most recently, \citet{zhang2026hyperagents} go beyond harness and further formulate the optimizer itself as an optimization target, opening a new paradigm for harness optimization.

\paragraph{Evaluating harness optimizers.}
The closest effort is a concurrent work by~\citet{ursekar2026vero}, which directly analyzes optimizer actions using \texttt{commit} histories. Yet, analyses are manual and limited to general statistics, \eg, number of prompt updates. Inspired by them and to close this gap, we present a low-cost, reliable way to directly evaluate an optimizer's ability.

\section{Why is Direct Evaluation Necessary for Harness Optimizers?}
\label{sec:preliminaries}

One may question: \textit{Isn't observing the end-improvement of target agents alone enough for evaluating harness optimizers? As long as the agent eventually improves, doesn't it mean that optimizers' mistakes are} (i) \textit{acceptable} or (ii) \textit{self-correcting?}
In this section, we argue against this assumption by analyzing real optimization trajectories, and discuss the need for direct evaluation.

\begin{wrapfigure}{r}{0.575\textwidth}
    \centering
    \vspace{-15pt}
    \includegraphics[width=0.575\textwidth]{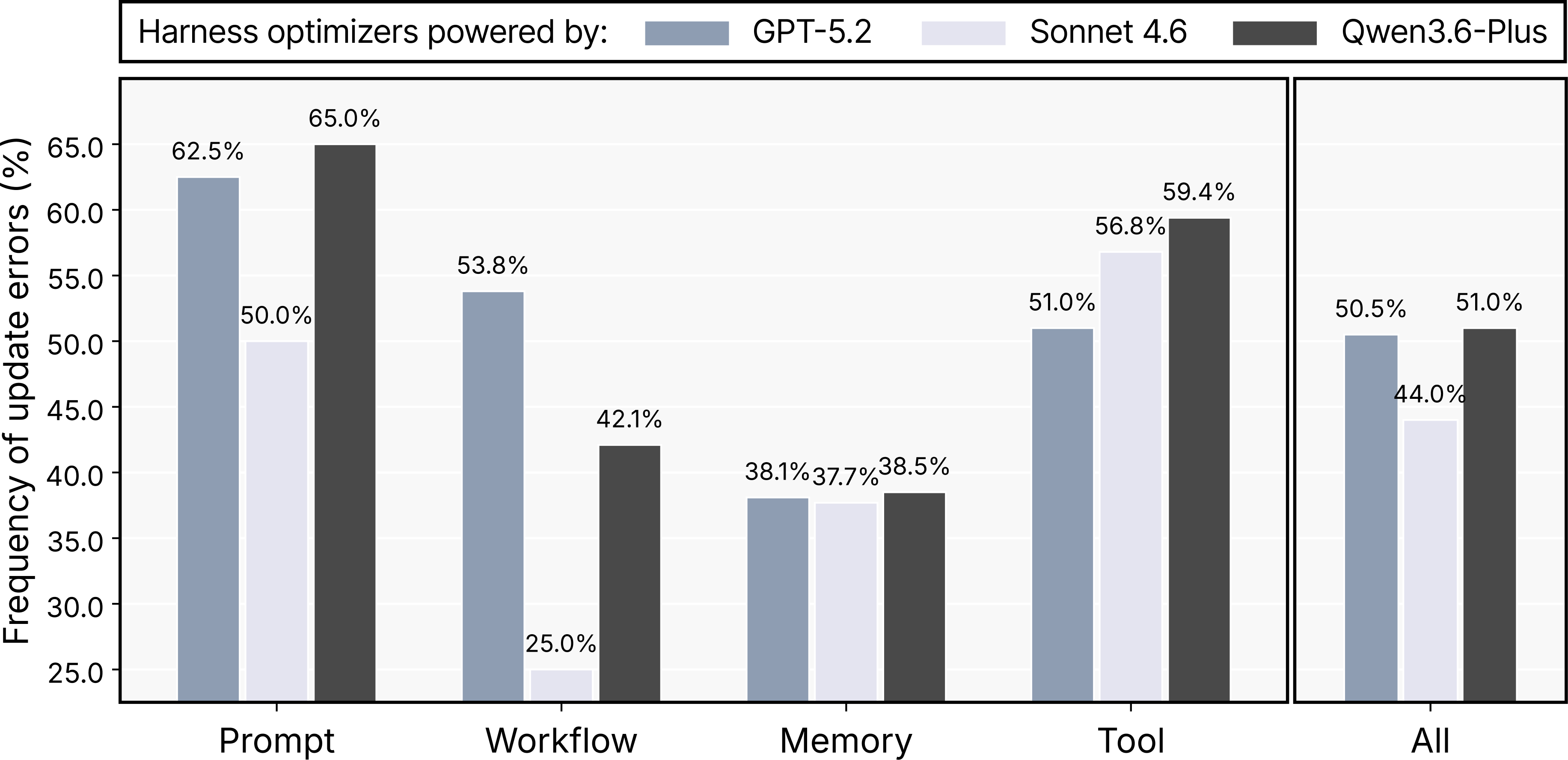}
    \vspace{-10pt}
    \caption{Frequency of erroneous updates over the optimization process, regarding each harness component.}
    \label{fig:errors}
    \vspace{-15pt}
\end{wrapfigure}

\paragraph{Analysis I: About half of the optimization steps are considered detrimental.}
Studies have reported that harness optimizers make mistakes during
the optimization process~\citep{ursekar2026vero, zhang2026hyperagents}.
To investigate the severity of this observation, we quantify it by examining real optimization trajectories (150 harnesses in total) across domains (Section~\ref{sec:evaluation_settings}),  where we use three optimizers based on ReCreate~\citep{hao2026recreate}. 
Figure~\ref{fig:errors} (right) shows that optimizers make erroneous updates in 44.0\textasciitilde51.0\% of the optimization steps. 
Verified by humans, these update actions are considered harmful for agent performance, when domain information and associated skills are taken into account (Appendix~\ref{app:preliminary}).
For example, we find a case where the system prompt asks a customer support agent to skip the greeting and start a session by calling a regex tool, resulting in 14 times of meaningless tool calls and eventually task failures.
This suggests that optimizers' erroneous steps should not be ignored when evaluating them.

\paragraph{Analysis II: Intermediate errors accumulate and appear in the final harness.} Then, we investigate whether these intermediate errors are self-correcting after the whole optimization process. We find that while some intermediate errors in \texttt{prompt} (17.8\%) do not persist in the final harness, 94.4\% of errors in other components remain not addressed. 
This suggests that harness optimizers' erroneous updates are often not temporary; they will eventually be part of the final harness and agent, and end-improvement evaluation will not surface them.

\begin{wraptable}{r}{0.60\textwidth}
    \centering
    \vspace{-10pt}
    \resizebox{0.6\textwidth}{!}{%
    \begin{tabular}{l||c|c|c|c|c}
    \toprule
          \cellcolor{gray!7}LLMs in the Optimizer & \cellcolor{gray!8}SWE-V & \cellcolor{gray!8}$\tau^2$-Bench & \cellcolor{gray!8}GAIA & \cellcolor{gray!8}Spider 2.0-lite & \textbf{\cellcolor{gray!8}Avg.} \\
    \midrule
        \textbf{GPT-5.2} & 0.636 & 0.545 & 0.545 & 0.545 & 0.568 \\
        \midrule
        \textbf{Sonnet 4.6} & 0.545 & 0.364 & 0.273 & 0.364 & 0.386 \\
        \midrule
        \textbf{Qwen3.6-Plus} & 0.545 & 0.455 & 0.364 & 0.364 &0.432 \\
    \bottomrule 
    \end{tabular}%
    }
    \vspace{2.5pt}
    \caption{Accuracy of harness optimizers in predicting whether the update actions they take on the harness improve or hinder agent performance.} 
    \vspace{-10pt}
    \label{tab:preliminary_binary}
\end{wraptable}

\paragraph{Analysis III: Harness optimizers lack awareness of the consequences of their own update actions.} 
While optimizers have access to agents' trajectories and performance, why do they still frequently take erroneous steps and leave them unaddressed? 
To get to the root cause, we investigate whether optimizers understand the effects of their own actions.
We instruct harness optimizers to perform binary classification of whether their update will improve or hinder agent SR, given consecutive harnesses from time $t$ and time $t+1$ and associated environmental information. Table~\ref{tab:preliminary_binary} shows that optimizers struggle to detect whether their update will benefit agents, yielding prediction accuracy close to (GPT-5.2 and Qwen3.6-Plus: 0.568 and 0.432) or much worse (Sonnet 4.6: 0.386) than random guessing (0.500).

\paragraph{Insights:} These findings reveal a fundamental problem: \textit{we currently have no means to assess whether a harness optimizer makes an informed update at each step.}
If optimization is well-directed, we expect errors to be infrequent and trivial, self-correcting, and detectable by optimizers themselves. Yet our analyses refute all three: performance-degrading errors occur in nearly half of optimization steps (Analysis I), persist to the final harness (Analysis II), and optimizers lack understanding of their actions (Analysis III). 
Together, we argue that \textbf{whether optimization steps are well-directed or simply driven by trial-and-error stays unclear}, and end-improvement evaluation cannot answer this. 
What is needed, thus, is an evaluation that directly quantifies optimizers' ability to prioritize effective updates over detrimental ones at the step level.\footnote{Implementation details Section~\ref{sec:preliminaries} and empirical error cases are presented in Appendix~\ref{app:preliminary}.}

\section{Priority Ranking to Evaluate Harness Optimizers}
\label{sec:method}
Directly evaluating how an optimizer performs in harness optimization is costly and time-consuming at scale. This is because it requires either human assessment of each version of the harnesses, or massive exploration of updatable components and their effects on agents.
To address this, we present \textit{priority ranking}, a simple yet efficient design to quantify optimizer ability at the step level.

\subsection{Background \& Design Rationale}
\label{ssec:background}
Before diving into priority ranking, we formally define harness optimization following~\citet{lee2026meta} and discuss our design rationale.
In harness optimization, an optimizer agent $\mathrm{OPT}$ should improve the performance of a target agent $A = (\phi, h)$ in task distribution $\mathcal{X}$ by iteratively updating the harness $h$, where $\phi$ is the LLM brain of $A$.
Specifically, at time step $t$, $\mathrm{OPT}$ updates $h_t$ and produces a new harness $h_{t+1}$ based on the optimization trajectory $\mathcal{H}_t$. 
Here, $\mathcal{H}_t = \{(h_i, \tau_i, r_i, S_{i})\}_{i=1}^{t}$, where $\tau_i$ is the execution trajectory of $A$ on a set of sampled tasks $X_i \sim \mathcal{X}$, and $r_i = r(\tau_i, X_i)$ is the performance of $A$ in $X_i$. The summary $S_{i}$ is written by $\mathrm{OPT}$, containing insights extracted from the behaviors of $A$ and $\mathrm{OPT}$.\footnote{The summary $S_{i}$ will be \texttt{empty} when $i=1$; Example summaries are provided in Appendix~\ref{app:other_details}.} Ideally, ${h}_{t+1}$ should enhance the performance of $A$ at step $t+1$:
\begin{equation}
    \label{eq:harness_opt}
    \textit{Harness Optimization at Each Step}:
    \mathcal{H}_t \;\xrightarrow[]{\mathrm{OPT}}\; h_{t+1}
\end{equation}
Here, $\mathrm{OPT}$ faces a fundamental decision at each step: \textbf{\textit{which part of the harness should it direct its update effort to?}} 
A strong optimizer must recognize which components need to be fixed or have room for improvement.
This prioritization, at its core, is a \textit{ranking problem over harness components}.   
Thus, framing step-level evaluation via ranking serves as a means to quantify optimizers' ability.
Moreover, this addresses the bottlenecks of direct optimizer evaluation:
(i) ranking needs only relative priorities among components, rather than absolute next harnesses that realistically do not exist; (ii) ranking is a non-iterative text generation task, excluding extensive rollouts of the optimizer and agent.

\subsection{Formulating Priority Ranking}
Now, we formulate \textit{priority ranking} as follows. Formally, a harness $h$ is the combination of $N$ components $\mathcal{C} = \{c_1, \ldots, c_N\}$. Following~\citet{hao2026recreate}, we formulate $\mathcal{C}$ as \texttt{prompt}, \texttt{tool}, \texttt{memory}, and \texttt{workflow}.\footnote{After systematically examining recent open-source, general-purpose agent harnesses and analyzing their design patterns, \citet{hao2026recreate} define an agent harness as the combination of these four modular components.} Given $\mathcal{H}_t = \{(h_i, \tau_i, r_i, S_{i})\}_{i=1}^{t}$ as file, $\mathrm{OPT}$ should interact with the file system\footnote{Both harness optimization (multi-step) and priority ranking (single-step) require multiple interactions with file systems within a single step.} and rank each component $c \in \mathcal{C}$ that makes up $h_t$, based on its potential to improve or harm agent performance at time step $t+1$:
\begin{equation}
    \textit{Priority Ranking}: \mathcal{H}_t \;\xrightarrow[]{\mathrm{OPT}}\; \bigl(c_{[1]} \succ \cdots \succ c_{[N]}\bigr)
\end{equation}
where $c_{[i]} \in \mathcal{C}$ denotes the component ranked $i$-th in terms of priority.

For example, consider a harness $h_t$. Currently, the system prompt lacks information about the environment, causing $A$ to repeatedly query irrelevant tools (revealed in $\tau_t$ and $S_t$). 
Meanwhile, the only memory-related issue in $h_t$ is that it is storing memories that contain the same problem-solving insights.
In this case, a strong $\mathrm{OPT}$ should assign $\texttt{prompt} \succ \texttt{memory}$, recognizing that updating the prompt will hold a larger potential to improve $A = (\phi, h_{t+1})$ than removing the redundant memory. This is because modifying \texttt{prompt} directly addresses the root cause of failure. In the following section, we introduce the dataset for this evaluation.

\section{The \shor~\textbf{\textsc{Shor}} Dataset}
\label{sec:our_dataset}
To enable \underline{s}tep-level evaluation of \underline{h}arness \underline{o}ptimizers via priority \underline{r}anking, we curate \textbf{\textsc{Shor}}, a collection of 182 human-verified scenarios of harness optimization steps (Figure~\ref{fig:dataset_generation}). Each is based on a base harness $h$ waiting to be updated and its priority ranking annotation $R$ for harness components.

\subsection{Harness Collection}
\label{ssec:collecting_base_harness}
To curate base harnesses in \textsc{Shor},
we start by collecting candidate harnesses
$\mathrm{h}_{\textit{Cand}} = \{{h}_T^{(j)}\}_{j=1}^{J}$ from real optimization trajectories.
We run a Claude Code~\citep{claude2026claudecode}-based Meta-Harness optimizer~\citep{lee2026meta} on tasks across domains and treat all versions of harnesses as $\mathrm{h}_{\textit{Cand}}$. Then, we examine each $h_T \in \mathrm{h}_{\textit{Cand}}$ to see if it contains an erroneous design. If so, we attach a \texttt{Flawed} label and provide an explanation. The error is determined based on domain information and skills, as in Section~\ref{sec:preliminaries}.
The resulting $\mathrm{h}_{\textit{Cand}}$ spans harnesses across domains and varying stages of the optimization, \ie, $T$ ranges from early to late optimization steps. $\mathrm{h}_{\textit{Cand}}$ is then annotated and filtered through the following steps.

\subsection{Priority Annotation}
\label{ssec:annotating}

We proceed to annotate the priority ranking $R$ for ${h}_T \in \mathrm{h}_{\textit{Cand}}$. Our protocol includes 3 steps: (i) Exploring different optimization directions by proposing various next harnesses from ${h}_T$. Each primarily targets the optimization of a component $c\in \mathcal{C}$; (ii) Assigning ranking to each $c$ based on how the corresponding harness affects agent SR; (iii) Filtering and verification to ensure data quality.

\paragraph{Step I: Exploring Component-targeted Optimization Directions.}
Based on ${h}_T$, we employ a group of coding agents (Section~\ref{sec:evaluation_settings}) as annotators $\mathbb{A}$ to explore optimization directions. 
Specifically, for each component $c \in \mathcal{C}$, each annotator $\alpha$ produces a set of next harness variants $h_{T+1}^{c,\alpha}$ by updating ${h}_T$ with a primary focus on $c$. This step yields a set of variants $\{{h}_{T+1}^{c, \alpha}\}_{c \in \mathcal{C}, \alpha \in \mathbb{A}}$, \ie, each ${h}_T$ has $|\mathcal{C}| \times |\mathbb{A}|$ variants of next harnesses.

\paragraph{Step II-1: Deriving Per-annotator Rankings.}
Next, for each annotator $\alpha$, we derive a component ranking for ${h}_T$. We start by running an agent $A$ on ${h}_T$ and each variant next harness ${h}_{T+1}^{c, \alpha}$ and measuring the SR change brought by each ${h}_{T+1}^{c, \alpha}$ :
\begin{equation}
\label{eq:per_annotator_ranking}
    \Delta\mathrm{SR}^{c, \alpha} =
r^A(\tau_{T+1}^{c, \alpha}, x_T) - r^A(\tau_T, x_T)
\end{equation}
where $\tau_{T+1}^{c, \alpha}$ denotes the execution trajectory of $A$ under ${h}_{T+1}^{c, \alpha}$. Then, each annotator $\alpha$ independently yields a ranking $R^{\alpha} = \left(c_{[1]} \succ \cdots \succ c_{[N]}\right)^{\alpha}$ by sorting $\mathcal{C}$ in descending order of $\Delta\mathrm{SR}^{c, \alpha}$. In other words, a component is ranked higher in priority if an update targeting it leads to a larger improvement. This yields a set of per-annotator rankings $\{R^{\alpha}\}_{\alpha \in \mathbb{A}}$ per ${h}_T$.

\paragraph{Step II-2: Reaching a Consensus Ranking.}
The next step is to reach a consensus ranking from per-annotator rankings. 
To ensure each data point has strong ranking signals, we keep only $h_{T}$ with high inter-annotator consistency before finalizing the ranking.
For that, we compute Kendall's W ($\omega$) among $\{R^{\alpha}\}_{\alpha \in \mathbb{A}}$ and discard ${h}_T$ if $\omega \leq \epsilon$.\footnote{We set $\epsilon = 0.500$ (Step II-2) and set $\delta = 5e-3$ (Step III).}
Then, for each remaining ${h}_T$, we finalize the priority ranking by aggregating per-annotator rankings based on the mean SR change regarding each $c \in \mathcal{C}$:
\begin{equation}
    \bar{\Delta}\mathrm{SR}^{c} = \frac{1}{|\mathbb{A}|}\sum_{\alpha \in \mathbb{A}} \Delta\mathrm{SR}^{c, \alpha}
\end{equation}
Similar to Step II-1, the final, consensus ranking label $R = \left(c_{[1]} \succ \cdots \succ c_{[N]}\right)$ is determined by sorting $\mathcal{C}$ in descending order of $\bar{\Delta}\mathrm{SR}^{c}$.

\paragraph{Step III: Ensuring Label Reliability via Performance Gap Filtering.}
To further ensure our labels provide strong enough ranking signals, we require that every adjacently ranked component in $R$ have a sufficiently distinct performance gap (\ie, larger than $\delta$). We discard instances failing the criterion:
\begin{equation}
    \bar{\Delta}\mathrm{SR}^{c_{[i]}} - \bar{\Delta}\mathrm{SR}^{c_{[i+1]}} > \delta, \quad \forall\, i \in [1, N-1]
\end{equation}

\paragraph{Human Verification and Finalization}. Finally, for every remaining instance, we manually verify the quality of ranking labels before including it as a base harness $h$ in \textsc{Shor}. 
Each instance $d$ in \textsc{Shor} is defined as follows:
\begin{equation}
    d^{(m)} = \left(
    \mathcal{H},\;
    T,\; 
    R,\;
    I
    \right), \quad \mathcal{H} = \{(h_i, \tau_i, r_i, S_{i})\}_{i=1}^{T}, \quad m \in [1, 182]
\end{equation}
where $T$ is the step at which $h$ is collected from a real trajectory, and $I$ contains metadata, harness type (\texttt{Flawed} or not), and artifacts from the annotation process. Instances (i) not included in \textsc{Shor} and (ii) attached with a \texttt{flawed} label are collected as a byproduct dataset, namely \texttt{SHOR-Flaw}.\footnote{All implementation details on dataset collection and dataset statistics are in Appendix~\ref{app:SHOR}.}

We discuss (i) the consistency between the ranking labels produced by our pipeline and those annotated by humans; and (ii) the consistency of the produced labels when the target agent used in Step II-1 is changed in Appendix~\ref{app:additional_results}.
\section{Main Experiments}
\subsection{Setups}
\label{sec:evaluation_settings}

\textbf{Environments.} For main experiments and \textsc{Shor}, we adopt 4 popular datasets across domains: (i) Spider 2.0-lite~\citep{lei2024spider} for text-to-SQL; (ii) $\tau^2$-Bench~\citep{barres2025tau} for customer service; (iii) GAIA~\citep{mialon2023gaia} for general intelligence; (iv) SWE-bench Verified~\citep{openai2024swe}. These are considered in-domain (ID).

\textbf{Out-of-domain environments.} For out-of-domain (OOD) evaluation, we apply (i) AppWorld~\citep{trivedi2024appworld} for interactive code generation; 
(ii) GPQA~\citep{rein2024gpqa} for scientific knowledge.

\textbf{Agent-based annotators for \textsc{Shor}.} In priority annotation, we have 3 SOTA coding agents work as optimizers to explore diverse optimization directions: Codex (GPT-5.3-Codex)~\citep{openai2025codex, openai2026gpt53codex}; Claude Code (Claude-Sonnet-4.6) (following Meta-Harness)~\citep{claude2026claudecode, claude2026sonnet46}; Gemini-CLI (Gemini-3-Pro)~\cite{geminicli2026geminicli, google2026gemini31pro}.

\textbf{Optimizers under evaluation.}
Including the above three optimizers, we further adopt mini-swe-agent (following ReCreate)~\citep{mini-swe-agent}, and OpenHands-CLI~\citep{wang2024openhands}.
More details on environments are in Appendix~\ref{app:datasets}, and details on optimizers/annotators can be found in Appendix~\ref{app:optimizers}.

\subsection{Harness Optimizers' Performance in Priority Ranking}
\label{sec:results}
\paragraph{Finding 1. Most optimizers struggle to rank components, and strong 
performance does not generalize across domains.}
Table~\ref{tab:optimizer-harness-results} reports optimizer performance on priority 
ranking via Acc@1 and normalized discounted cumulative gain (NDCG).\footnote{Acc@1 measures the percentage of times the top-ranked component is the one labeled as the top priority. NDCG evaluates the quality of the entire ranking with regard to the ranking label.} Overall, optimizers exhibit limited ability to identify harness components that deserve prioritized update efforts, with OpenHands-CLI (DeepSeek-V4-Pro) leading in Acc@1 (0.305) and Claude Code (Sonnet 4.6) achieving the 
highest NDCG of 0.793. 
Within the same harness, optimizers with stronger base LLMs generally are better at ranking, except for mini-swe-agent. Notably, ranking performance is 
inconsistent across domains: optimizers excelling in SWE-V, \eg, OpenHands-CLI (DeepSeek-V4-Pro) and Gemini-CLI (Gemini 3 Pro), struggle in $\tau^2$-Bench,  and Claude Code (Sonnet 4.6) yields relatively high performance across all domains except for Spider. This suggests that no single optimizer, not even those commonly considered powerful, masters component prioritization across all optimization scenarios.

\begin{table*}[t]
\centering
\vspace{5em}
\small
\resizebox{1\linewidth}{!}{
\setlength{\tabcolsep}{3.5pt}
\begin{tabular}{l|l|cccccccc|cc}
\toprule
\multicolumn{2}{c}{Harness Optimizer ($\textrm{OPT}$)}
& \multicolumn{2}{c}{GAIA}
& \multicolumn{2}{c}{Spider 2.0-lite}
& \multicolumn{2}{c}{SWE-V}
& \multicolumn{2}{c}{$\tau^2$-Bench}
& \multicolumn{2}{c}{\textbf{Average}} \\
\cmidrule(lr){3-4}
\cmidrule(lr){5-6}
\cmidrule(lr){7-8}
\cmidrule(lr){9-10}
\cmidrule(lr){11-12}
$\textrm{OPT}$'s Harness & $\textrm{OPT}$'s Base LLMs & Acc@1 & NDCG
& Acc@1 & NDCG
& Acc@1 & NDCG
& Acc@1 & NDCG
& \textbf{Acc@1} & \textbf{NDCG} \\
\midrule
\multirow{3}{*}{mini-swe-agent}
& GLM-5.1            & \cellcolor[HTML]{8E9DB2}{0.303} & \cellcolor[HTML]{AEB9C8}{0.796} & \cellcolor[HTML]{F8F9FB}{0.186} & \cellcolor[HTML]{F8F9FB}{0.745} & \cellcolor[HTML]{D7DCE4}{0.265} & \cellcolor[HTML]{E4E8ED}{0.788} & \cellcolor[HTML]{AEB9C8}{0.267} & \cellcolor[HTML]{D7DCE4}{0.766} & \cellcolor[HTML]{AEB9C8}{0.255} & \cellcolor[HTML]{D7DCE4}{0.774} \\
& GPT-5.2            & \cellcolor[HTML]{E5E8ED}{0.212} & \cellcolor[HTML]{E4E8ED}{0.764} & \cellcolor[HTML]{AEB9C8}{0.279} & \cellcolor[HTML]{D7DCE4}{0.772} & \cellcolor[HTML]{F8F9FB}{0.206} & \cellcolor[HTML]{F8F9FB}{0.779} & \cellcolor[HTML]{F8F9FB}{0.167} & \cellcolor[HTML]{D7DCE4}{0.767} & \cellcolor[HTML]{F8F9FB}{0.216} & \cellcolor[HTML]{E4E8ED}{0.771} \\
& DeepSeek-V4-Pro    & \cellcolor[HTML]{F8F9FB}{0.182} & \cellcolor[HTML]{F8F9FB}{0.756} & \cellcolor[HTML]{E5E8ED}{0.233} & \cellcolor[HTML]{E4E8ED}{0.767} & \cellcolor[HTML]{9DAABD}{0.353} & \cellcolor[HTML]{9DAABD}{0.824} & \cellcolor[HTML]{D7DCE4}{0.233} & \cellcolor[HTML]{E4E8ED}{0.757} & \cellcolor[HTML]{D7DCE4}{0.250} & \cellcolor[HTML]{D7DCE4}{0.776} \\
\midrule
\multirow{3}{*}{OpenHands-CLI}
& GLM-5.1            & \cellcolor[HTML]{F8F9FB}{0.182} & \cellcolor[HTML]{E4E8ED}{0.764} & \cellcolor[HTML]{E5E8ED}{0.233} & \cellcolor[HTML]{E4E8ED}{0.771} & \cellcolor[HTML]{E5E8ED}{0.235} & \cellcolor[HTML]{E4E8ED}{0.796} & \cellcolor[HTML]{9DAABD}{0.300} & \cellcolor[HTML]{AEB9C8}{0.787} & \cellcolor[HTML]{E5E8ED}{0.237} & \cellcolor[HTML]{AEB9C8}{0.779} \\
& GPT-5.2            & \cellcolor[HTML]{E5E8ED}{0.212} & \cellcolor[HTML]{F8F9FB}{0.756} & \cellcolor[HTML]{9DAABD}{0.349} & \cellcolor[HTML]{AEB9C8}{0.796} & \cellcolor[HTML]{D7DCE4}{0.265} & \cellcolor[HTML]{E4E8ED}{0.795} & \cellcolor[HTML]{F8F9FB}{0.133} & \cellcolor[HTML]{F8F9FB}{0.726} & \cellcolor[HTML]{E5E8ED}{0.240} & \cellcolor[HTML]{F8F9FB}{0.768} \\
& DeepSeek-V4-Pro    & \cellcolor[HTML]{AEB9C8}{0.273} & \cellcolor[HTML]{E4E8ED}{0.764} & \cellcolor[HTML]{8E9DB2}{0.372} & \cellcolor[HTML]{AEB9C8}{0.800} & \cellcolor[HTML]{8E9DB2}{0.441} & \cellcolor[HTML]{8E9DB2}{0.839} & \cellcolor[HTML]{F8F9FB}{0.133} & \cellcolor[HTML]{F8F9FB}{0.736} & \cellcolor[HTML]{8E9DB2}{0.305} & \cellcolor[HTML]{AEB9C8}{0.785} \\
\midrule
\multirow{2}{*}{Gemini-CLI}
& Gemini 3 Flash     & \cellcolor[HTML]{AEB9C8}{0.273} & \cellcolor[HTML]{AEB9C8}{0.784} & \cellcolor[HTML]{F8F9FB}{0.209} & \cellcolor[HTML]{E4E8ED}{0.766} & \cellcolor[HTML]{D7DCE4}{0.294} & \cellcolor[HTML]{AEB9C8}{0.801} & \cellcolor[HTML]{F8F9FB}{0.100} & \cellcolor[HTML]{F8F9FB}{0.728} & \cellcolor[HTML]{E5E8ED}{0.219} & \cellcolor[HTML]{E4E8ED}{0.770} \\
& Gemini 3 Pro       & \cellcolor[HTML]{E5E8ED}{0.212} & \cellcolor[HTML]{E4E8ED}{0.766} & \cellcolor[HTML]{AEB9C8}{0.302} & \cellcolor[HTML]{AEB9C8}{0.793} & \cellcolor[HTML]{9DAABD}{0.412} & \cellcolor[HTML]{9DAABD}{0.830} & \cellcolor[HTML]{E5E8ED}{0.233} & \cellcolor[HTML]{E4E8ED}{0.759} & \cellcolor[HTML]{9DAABD}{0.290} & \cellcolor[HTML]{9DAABD}{0.787} \\
\midrule
\multirow{2}{*}{Claude Code}
& Haiku 4.5   & \cellcolor[HTML]{E5E8ED}{0.212} & \cellcolor[HTML]{AEB9C8}{0.779} & \cellcolor[HTML]{D7DCE4}{0.256} & \cellcolor[HTML]{E4E8ED}{0.771} & \cellcolor[HTML]{E5E8ED}{0.235} & \cellcolor[HTML]{E4E8ED}{0.786} & \cellcolor[HTML]{F8F9FB}{0.167} & \cellcolor[HTML]{F8F9FB}{0.745} & \cellcolor[HTML]{F8F9FB}{0.217} & \cellcolor[HTML]{E4E8ED}{0.770} \\
& Sonnet 4.6  & \cellcolor[HTML]{9DAABD}{0.303} & \cellcolor[HTML]{8E9DB2}{0.813} & \cellcolor[HTML]{E5E8ED}{0.233} & \cellcolor[HTML]{F8F9FB}{0.763} & \cellcolor[HTML]{AEB9C8}{0.294} & \cellcolor[HTML]{AEB9C8}{0.808} & \cellcolor[HTML]{8E9DB2}{0.333} & \cellcolor[HTML]{8E9DB2}{0.789} & \cellcolor[HTML]{AEB9C8}{0.291} & \cellcolor[HTML]{8E9DB2}{0.793} \\
\midrule
\multirow{2}{*}{Codex}
& GPT-5.2            & \cellcolor[HTML]{F8F9FB}{0.152} & \cellcolor[HTML]{F8F9FB}{0.742} & \cellcolor[HTML]{D7DCE4}{0.256} & \cellcolor[HTML]{E4E8ED}{0.767} & \cellcolor[HTML]{F8F9FB}{0.176} & \cellcolor[HTML]{E4E8ED}{0.788} & \cellcolor[HTML]{9DAABD}{0.300} & \cellcolor[HTML]{AEB9C8}{0.780} & \cellcolor[HTML]{F8F9FB}{0.221} & \cellcolor[HTML]{F8F9FB}{0.769} \\
& GPT-5.5            & \cellcolor[HTML]{E5E8ED}{0.242} & \cellcolor[HTML]{AEB9C8}{0.792} & \cellcolor[HTML]{F8F9FB}{0.209} & \cellcolor[HTML]{E4E8ED}{0.766} & \cellcolor[HTML]{E5E8ED}{0.235} & \cellcolor[HTML]{E4E8ED}{0.788} & \cellcolor[HTML]{F8F9FB}{0.200} & \cellcolor[HTML]{F8F9FB}{0.741} & \cellcolor[HTML]{F8F9FB}{0.222} & \cellcolor[HTML]{E4E8ED}{0.772} \\
\midrule
\midrule
\multicolumn{2}{l|}{Random}
& 0.250 & 0.780
& 0.250 & 0.780
& 0.250 & 0.780
& 0.250 & 0.780
& 0.250 & 0.780 \\
\midrule
\multicolumn{2}{l|}{Human Performance (Oracle)}
& 0.296 & 0.785
& 0.185 & 0.752
& 0.370 & 0.815
& 0.333 & 0.797
& 0.296 & 0.787 \\

\bottomrule
\end{tabular}
}
\vspace{5pt}
\caption{Harness optimizers' performance in priority ranking.}
% \vspace{-1em}
\label{tab:optimizer-harness-results}
\end{table*}

\paragraph{Finding 2. Good agent harnesses are not necessarily good optimizer 
harnesses.}
When the LLM brain is fixed and the optimizer harness is varied, a counterintuitive 
pattern emerges. GPT-5.2 paired with OpenHands-CLI outperforms GPT-5.2 
paired with Codex~\textemdash~widely regarded as a stronger agent harness from the same company~\textemdash~in 3 out of 4 domains, with pronounced gaps in 
Spider~2.0-lite (+9.3\%p) and SWE-V (+8.9\%p). This indicates that harness designs that excel on mainstream 
agentic benchmarks do not necessarily transfer to the harness optimization setting: what 
constitutes a \textit{good} optimizer harness may diverge from what the research 
community currently optimizes for. This finding is partially supported by 
\citet{zhang2026hyperagents}, who show that explicitly treating the optimizer's own 
harness as an optimization target yields better optimization performance, proving the non-trivial role of dedicated harness design for harness optimizers.

\begin{wraptable}{r}{0.425\textwidth}
    \centering
    \vspace{-1em}
    \resizebox{0.4\textwidth}{!}{%
    \begin{tabular}{l||c|c}
    \toprule
     &  Cost (\$) & Time (min)  \\
    \midrule
        End-SR Observation & 524.3 & 1089.8\\
        \midrule
        Priority Ranking & \textbf{63.7} & \textbf{63.2} \\
    \bottomrule 
    \end{tabular}%
    }
    \vspace{5pt}
    \caption{Cost and elapsed time of end-SR observation \& priority ranking.} 
    \vspace{-1.25em}
    \label{tab:time_and_cost}
\end{wraptable}

\paragraph{Finding 3. Priority ranking is 8$\times$ cheaper and 17$\times$ faster than end-improvement observation.} 
Compared to end-improvement observation, which requires running the full harness optimization, priority ranking via  \textsc{Shor} exhibits obvious cost and time efficiency (Table~\ref{tab:time_and_cost}; avg. of optimizers above). Implementation details are in Appendix~\ref{app:cost_and_time}.

\paragraph{Does all of this matter?} Priority ranking provides a low-cost means to quantify optimizers' ability at the step level. However, this raises a question: Why does this evaluation matter? In the next section, we investigate whether priority ranking can serve as a reliable proxy for anticipating optimizers' performance in real optimization settings.

\subsection{Priority Ranking’s Correlation with Full Harness Optimization}
\label{ssec:result_corr_harness_optimization}

\begin{figure}[t]
    \centering
    \includegraphics[width=1\linewidth]{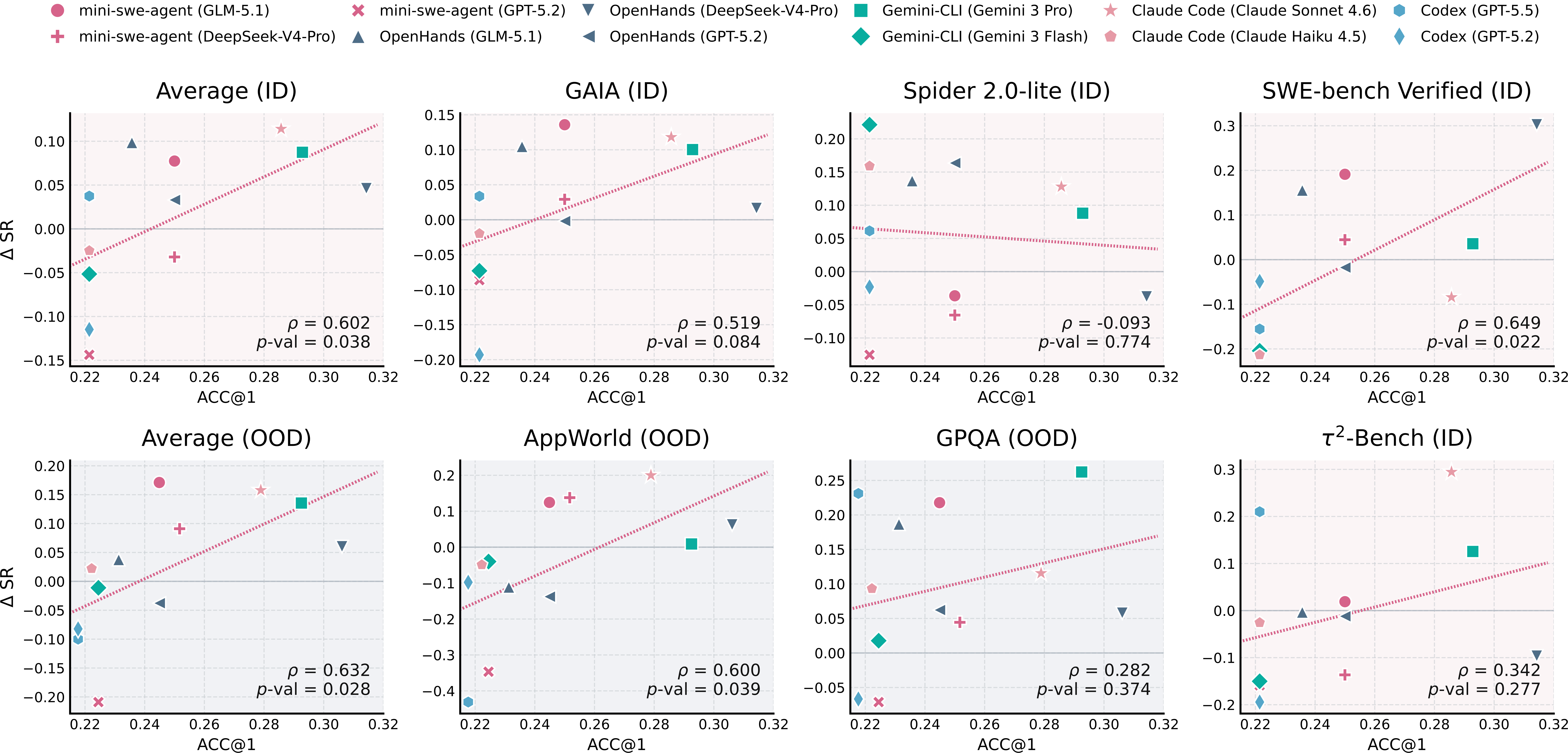}
    \vspace{0pt}
    \caption{Correlation between priority ranking and optimizer ability to improve target agents' SR in harness optimization. The harness optimization is run for 10 iterations. We report the average results of 5 runs. We use mini-swe-agent (gpt-5-mini) as the target agent. We report the analysis plotted with NDCG in Figure~\ref{fig:in_correlation_ndcg}. For generalizability, we further provide results from mini-swe-agent (Qwen-3.5-flash) and other additional results in Appendix~\ref{app:additional_results}.}
    \label{fig:in_correlation}
    % \vspace{-1.5em}
\end{figure}

\paragraph{Finding 4. Optimizers' performance in priority ranking correlates with the ability to improve target agents via harness optimization.} Figure~\ref{fig:in_correlation} plots each optimizer's Acc@1 against its target agents' SR change under actual harness optimization. A significant positive correlation (Pearson $\rho = 0.602$; p-value $= 0.038$) indicates that optimizers that better identify high-priority components tend to achieve greater target agent improvement in practice. The exception of Spider 2.0-lite suggests that harness optimization does not hinge solely on the ability to prioritize update efforts, \textit{but also on the ability to actually carry out the optimization actions}. We investigate this ability in Section~\ref{ssec:oracle_component_error_recover}.

\paragraph{Finding 5. The correlation generalizes to OOD settings.}
Moreover, when evaluating on two OOD environments: AppWorld ($\rho = 0.600$) and GPQA ($\rho = 0.282$), we observe positive correlation comparable to in-domain environments. This suggests that priority ranking retains predictive value in OOD settings. One way to explain this generalizability is that priority ranking, by design, focuses on relative optimization priorities rather than domain-specific solutions.

\begin{wrapfigure}{t}{0.4\textwidth}
    \centering
    \vspace{-10pt}
\includegraphics[width=0.4\textwidth]{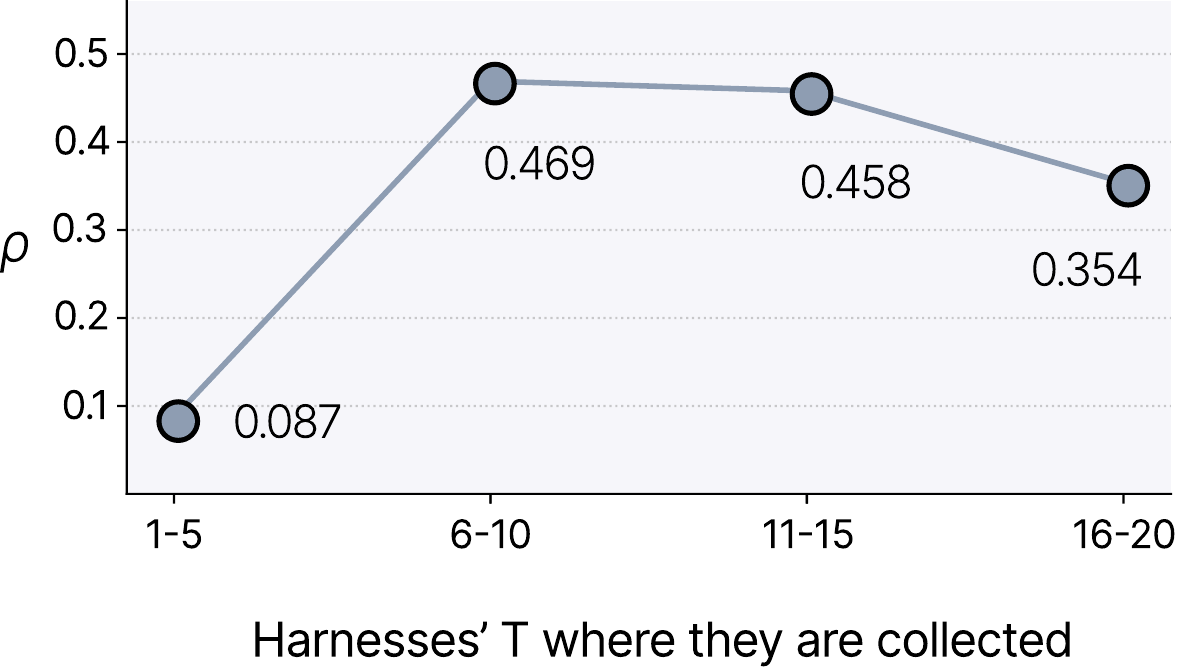}
    \vspace{-5pt}
    \caption{Correlation $\rho$ across time step intervals of base harnesses.}
    \label{fig:time_step}
    \vspace{-10pt}
\end{wrapfigure}

\paragraph{Finding 6. Ranking ability in mid-stage harnesses yields the strongest correlation.}
Figure~\ref{fig:time_step} plots the correlation, broken down by the 
time step $T$ at which the base harness in \textsc{SHOR} was collected from the optimization trajectory. The correlation peaks in the mid-stage window ($T \in [6, 10]$) and remains relatively strong through $T \in [11, 15]$, before declining in later stages. 
This mid-stage advantage can be conjectured through harness maturity. In 
early stages, harnesses are far from any optimum, and all components 
remain highly updatable, making relative priorities harder to differentiate. In late 
stages, harness design converges, weakening the ranking signal. Mid-stage harnesses strike a balance: enough past experience to surface meaningful component-level differences, yet enough room for improvement to amplify them.

Overall, this general trend of positive correlations justifies our ranking design: In actual harness optimization, the optimizer must understand the relationship between components' current functional state and expected agent performance. Priority ranking tests the same ability, just in isolation. While this is being said, we note that priority ranking should be treated as a \textit{necessary} rather than
\textit{sufficient} condition for evaluating harness optimization. Graphs plotted via NDCG and implementation details in this section are in Figure~\ref{fig:in_correlation_ndcg} and Appendix~\ref{app:main_exp}.

\begin{table*}[t]
\centering
\vspace{5em}
\small
\resizebox{1\linewidth}{!}{
\setlength{\tabcolsep}{3.5pt}
\begin{tabular}{l|l|cccccccc|cc}
\toprule
\multicolumn{2}{c|}{Harness Optimizer ($\textrm{OPT}$)}
& \multicolumn{2}{c}{GAIA}
& \multicolumn{2}{c}{Spider 2.0-lite}
& \multicolumn{2}{c}{SWE-V}
& \multicolumn{2}{c|}{$\tau^2$-Bench}
& \multicolumn{2}{c}{\textbf{Average}} \\
\cmidrule(lr){3-4} 
\cmidrule(lr){5-6} 
\cmidrule(lr){7-8} 
\cmidrule(lr){9-10} 
\cmidrule(lr){11-12}

$\textrm{OPT}$'s Harness & $\textrm{OPT}$'s Base LLMs
& \XSolidBrush & \Checkmark 
& \XSolidBrush & \Checkmark 
& \XSolidBrush & \Checkmark 
& \XSolidBrush & \Checkmark 
& \XSolidBrush & \Checkmark \\
\midrule

Gemini-CLI
& Gemini 3 Pro
& \cellcolor[HTML]{BFC8DF}{0.314}
& \cellcolor[HTML]{95A2CB}{0.657}
& \cellcolor[HTML]{ABB6D6}{0.346}
& \cellcolor[HTML]{7F8EBF}{0.704}
& \cellcolor[HTML]{ABB6D6}{0.400}
& \cellcolor[HTML]{ABB6D6}{0.500}
& \cellcolor[HTML]{ABB6D6}{0.346}
& \cellcolor[HTML]{7F8EBF}{0.815}
& \cellcolor[HTML]{ABB6D6}{0.350}
& \cellcolor[HTML]{95A2CB}{0.664} \\
\midrule

Claude Code
& Sonnet 4.6
& \cellcolor[HTML]{D2D8E8}{0.171}
& \cellcolor[HTML]{ABB6D6}{0.343}
& \cellcolor[HTML]{D2D8E8}{0.222}
& \cellcolor[HTML]{ABB6D6}{0.333}
& \cellcolor[HTML]{E1E5F0}{0.100}
& \cellcolor[HTML]{ABB6D6}{0.533}
& \cellcolor[HTML]{D2D8E8}{0.185}
& \cellcolor[HTML]{ABB6D6}{0.481}
& \cellcolor[HTML]{D2D8E8}{0.168}
& \cellcolor[HTML]{ABB6D6}{0.420} \\
\midrule

Codex
& GPT-5.5
& \cellcolor[HTML]{BFC8DF}{0.286}
& \cellcolor[HTML]{ABB6D6}{0.486}
& \cellcolor[HTML]{E1E5F0}{0.111}
& \cellcolor[HTML]{ABB6D6}{0.370}
& \cellcolor[HTML]{BFC8DF}{0.267}
& \cellcolor[HTML]{ABB6D6}{0.467}
& \cellcolor[HTML]{BFC8DF}{0.296}
& \cellcolor[HTML]{BFC8DF}{0.296}
& \cellcolor[HTML]{BFC8DF}{0.244}
& \cellcolor[HTML]{ABB6D6}{0.412} \\
\bottomrule
\end{tabular}
}
\vspace{5pt}
\caption{Human evaluation on optimizer performance (resolved rate) to resolve flawed harnesses that contain errors (\ie, the \texttt{SHOR-Flaw} dataset). \XSolidBrush and \Checkmark denote whether the location (\ie, which component) of serious flaws is provided. We further report the recovery performance of optimizers when they are further granted access to the human-annotated explanation of errors (Appendix~\ref{app:additional_results}).}
\vspace{-0.5em}
\label{tab:optimizer-harness-resolve}
\end{table*}

\section{Priority Ranking in Practice}
\label{sec:discussion}
\subsection{Priority Ranking as an Actionable Insight for Optimizer Development}
\label{ssec:oracle_component_error_recover}
Besides serving as a means to evaluate harness optimizers, \textit{does the concept of priority ranking provide further value?} Drawing inspiration from Section~\ref{sec:preliminaries}, we investigate the answer to this question by testing optimizers' ability to resolve erroneous harnesses under two conditions: (i) given a flawed harness; (ii) given a flawed harness, as well as information on which component is having an error. Table~\ref{tab:optimizer-harness-resolve} shows that while optimizers can not effectively resolve flawed harnesses, providing them with oracle information on which component possesses serious flaws largely improves resolving performance by 17$\sim$51\%p on average. This improvement is even more profound when looking at specific domains such as Spider (+72.0\%p in Gemini-CLI with Gemini 3 Pro). This shows that when the optimization priority is made explicit, optimizers are better able to carry out the target update.

Combined with our earlier findings (Finding 1), we conjecture that the ability to fix erroneous harness components exists in current SOTA coding agents (\ie, optimizers); what is lacking is the diagnostic procedure and ability to identify where to act, \ie, identify which part of the harness is in need of prioritized optimization efforts.
Consequently, an optimizer's performance in priority ranking is not merely an abstract evaluation score but a direct criterion for whether it can self-diagnose and direct its optimization effort effectively, aligning with Findings 4 and 5.

These findings point out a direction for building stronger harness optimizers: explicitly enhancing their ability to recognize parts of harnesses that require prioritized optimization effort. For instance, we may develop and include a dedicated priority prediction module into optimizers' workflow, decoupling priority identification from update actions so that the optimizer first identifies the component requiring the most prioritized attention before committing to an update. 
Importantly, the annotation pipeline for \textsc{Shor} provides a replicable procedure for acquiring ranking signals for such a module. 

We also report the recovery performance of optimizers when they are further granted access to the human-annotated explanation of errors (Table~\ref{tab:optimizer-harness-resolve-full-oracle}).

\subsection{Implications for Industrial Deployment}
As harness optimization transitions from research prototypes to production systems~\citep{Anthropic_2026, tang2025autoagent},
practitioners face a set of challenges that end-improvement evaluation alone is insufficient to address. We discuss the implications of our findings for industrial deployment around two topics: risks that remain invisible without direct evaluation, and practical opportunities that priority ranking opens up.

\textbf{Silent risks of harness optimizers in production.}
We reveal that harness optimizers' update steps are often detrimental to agent performance.
For enterprises deploying harness optimizers in high-stakes settings, \eg, clinical diagnosis, this means that a large portion of automated harness updates may silently hinder agent behavior, without any mechanism to surface them. Moreover, an optimizer validated on one business scenario offers no safety guarantees when redeployed to other domains.

\textbf{Priority ranking in production chains.}
On the other hand, priority ranking may synergize well with the enterprise agent lifecycle.
Before deployment, priority ranking functions as a low-cost screening
mechanism for optimizer selection: rather than running full optimization loops
to compare candidates~\textemdash~which we show is \(8\times\) more expensive
and \(17\times\) slower.
During operation, the single-step nature of priority ranking makes it suitable for integration into CI/CD pipelines as an automated quality gate, identifying harness updates that do not address predicted prioritization before they reach production agents. Although we acknowledge that this requires the optimizers to achieve acceptable performance in priority ranking, which even SOTA agentic frameworks struggle with.

\section{{Discussions and Limitations}}
\label{app:limitations}

\textbf{Simplified formulation.}
Following~\citet{hao2026recreate}, our harness formulation uses 4 categories of harness components, whereas the full harness may be decomposed into finer-grained units.
This abstraction may underestimate sensitivity to changes at a finer level. 
Extending the harness to a more fine-grained decomposition is a natural direction for future work.

\textbf{Single target agent.} 
Our study focuses on a single target agent throughout evaluation. Thus, our findings may not transfer to multi-agent systems, which involve multiple agents collaborating, competing, or sharing harness components.
Nevertheless, our design is not inherently restricted to single-agent systems: priority ranking requires only a defined set of harness components and measurable performance for annotation. For orchestrator-centered multi-agent systems, individual agents may, for example, be formulated as tools within the orchestrator's harness; more complex systems may require expanding the component space to capture factors such as interaction topology.

% \paragraph{Safety and Emergent Risks.} We do not explicitly address the safety implications of automated harness optimization. As discussed in the Broader Impact section (Appendix~\ref{app:societal_impact}), self-evolving agent systems may exhibit emergent and unintended behaviors~\citep{shao2025your}, yet our current framework provides no mechanism for detecting or mitigating such risks during the optimization process. 
% Future work may incorporate safety-aware guardrails, such as monitoring optimizer reasoning or optimization trajectories, to flag potentially harmful updates or misbehaviors stemming from prior update errors.

\textbf{Annotator composition.}
Ranking labels in \textsc{Shor} are derived from three SOTA coding agents (\ie, Codex, Claude Code, and Gemini-CLI), whose rankings are aggregated mechanically based on the measured SR changes of the target agent, inter-annotator consistency, performance gap filtering, and human verification. 
While this protocol enables reliable consensus labels, the resulting priorities are bounded by the updates these agents can produce and may not reflect the theoretical maximum improvement achievable. 
Eliciting labels from a more diverse annotator pool may surface aspects of harness optimization, which is a natural direction for future work.

\textbf{Construction cost.}
While \textsc{Shor} enables substantially cheaper evaluation, its annotation pipeline itself requires non-trivial one-time cost. 
Under our current setup, constructing \textsc{Shor} costs approximately \$4,000 per domain, although this cost varies with factors such as the number of annotators and filtering criteria. 
Our current filtering retains only approximately 30\% of candidate labels; relaxing these criteria could reduce construction cost at the expense of label reliability.

\section{Conclusion}
We explore the direct evaluation of harness optimizers, an aspect largely overlooked despite the growing trend of harness optimization. 
Our evaluation design, \textit{priority ranking}, quantifies optimizers' ability to make effective harness updates in isolation,
without expensive rollouts or human assessment. Its efficiency, combined with
a significant correlation to actual optimization ability ($\rho = 0.602$ and $0.632$), establishes priority ranking as a reliable and necessary evaluation procedure for harness optimizers, in addition to the common black-box end-improvement observation. 
Also, we demonstrate that priority ranking has value beyond serving as an evaluation design; it provides \textit{actionable insights} for developing better harness optimizers and addressing deployment-time safety risks.

\section*{Acknowledgement}
This work was mainly supported by Samsung Electronics Co., Ltd (IO251218-14819-01), with additional support from the following government-funded projects: This work was supported by Institute for Information \& Communications Technology Planning \& Evaluation (IITP) grant funded by the Korea government (MSIT) (No.RS-2026-25585074, Development of Agentic AI Technology for Enterprise-wide Business Innovation Based on Goal-Oriented Long-term Context Reasoning: 70\%); This work was supported by Institute of Information \& communications Technology Planning \& Evaluation (IITP) grant funded by the Korea government (MSIT) (No.RS-2024-00457882, National AI Research Lab Project: 20\%); This work was supported by Institute of Information \& Communications Technology Planning \& Evaluation (IITP) grant funded by the Korean government (MSIT) (No.RS-2020-II201361, Artificial Intelligence Graduate School Program (Yonsei University): 10\%). 
None of the authors have financial relationships with entities that could potentially be perceived to influence what we wrote in this work.
Jinyoung Yeo is the corresponding author.

\bibliographystyle{unsrtnat}
\bibliography{main}

\newpage
\appendix
\section{Appendix Contents}
\begin{itemize}
    \item Details on Analyses in Section~\ref{sec:preliminaries}: Appendix~\ref{app:preliminary}
    \item Details on \textsc{Shor}: Appendix~\ref{app:SHOR} 
    \item Details on Harness Optimizers and Agents Applied in This Study: Appendix~\ref{app:optimizers}
    \item Details on Domains and Datasets: Appendix~\ref{app:datasets}
    \item Details on Experiments: Appendix~\ref{app:main_exp}
    \item Other Details: Appendix~\ref{app:other_details}
    \item Additional Results: Appendix~\ref{app:additional_results}
    \item Computing Resources: Appendix~\ref{app:computing_resources}
    \item Prompts:
    Appendix~\ref{app:prompts}
    \item Societal Impact and Potential Harmful Consequences: Appendix~\ref{app:societal_impact}
    \item License: Appendix~\ref{app:license}
\end{itemize}

\section{Details on Analyses in Section~\ref{sec:preliminaries}}
\label{app:preliminary}

\subsection{Analysis I and II}
\label{app:preliminary_1_2}
\paragraph{Acquiring Real Optimization Trajectories.}
Following ReCreate~\citep{hao2026recreate}, we instantiate three optimizers with GPT-5.2, Claude Sonnet 4.6, and Qwen3.6-Plus    as the LLM brain for the optimizer agent, which updates the target agent's harness across four components: a system prompt, a workflow, a set of tools, and a memory module.
For the target agent $A$, we use gpt-5-mini as the LLM brain $\phi$ and adopt mini-swe-agent as the initial harness $h_0$.
Each optimizer runs for 10 iterations, where the target agent is evaluated on 4 randomly sampled validation tasks per iteration, and final performance is measured on a held-out test set of 40 instances.
All tasks are executed within Docker sandboxes, and each iteration is logged as a versioned harness folder with diffs and statistics to enable reproducible comparisons.
In total, this yields 15 optimization trajectories including 150 intermediate/final harnesses.

\paragraph{Error Annotation.}
We recruit four human annotators with relevant research experience, each assigned to one of the four domains, to manually examine the collected optimization trajectories. 
For each time step $t$, annotators are presented with the optimization trajectory $\mathcal{H}_t$ and the harness pair $(h_t, h_{t+1})$, and asked to (i) identify any updates that appear erroneous or undesirable from a human perspective, and (ii) annotate from which harness component the error originates.

\paragraph{Error Types.}
The identified erroneous updates are then aggregated and reviewed by the authors to identify recurring patterns. 
Through manual clustering, we consolidate the collected errors into eight categories, which we present in Table~\ref{tab:error-taxonomy}, along with their frequency and relative proportion across all identified updates.
Representative cases are provided in Table~\ref{tab:error-cases}.

\begin{table*}[h]
\centering
\resizebox{0.9\linewidth}{!}{%
\begin{tabular}{p{0.3\linewidth} p{0.6\linewidth}}
\toprule
\textbf{Error Type} & \textbf{Description} \\
\midrule
Redundant Duplication & The optimizer creates tools or memory components that already exist in the harness, introducing unnecessary redundancy that undermines system consistency and increases decision-making overhead. \\
\midrule
Hardcoding & The optimizer embeds task-specific values observed during optimization directly into the harness without generalization, limiting reusability across varying input distributions. \\
\midrule
Task-specific Addition & The optimizer introduces tools, memory, or instructions that are only valid for a narrow subset of tasks, which may be ineffective or cause errors when applied to other task instances. \\
\midrule
Hallucination & The optimizer references tools, memory, or information that do not exist in the execution environment, leading to runtime failures and degraded system reliability. \\
\midrule
Overengineering & The optimizer implements simple logic or responses as unnecessary tools or memory components, or continuously appends content without pruning, increasing structural complexity and disrupting the agent's reasoning flow. \\
\midrule
Direct Performance-degrading Update & The optimizer makes structural or prompt-level changes that directly reduce task performance, such as removing established response formats that are critical to agent behavior. \\
\midrule
Overgeneralized Heuristic & The optimizer maps diverse phenomena to a single cause or action, constraining the agent's reasoning by prematurely closing off alternative explanations or strategies. \\
\midrule
Safety Violation & The optimizer introduces changes that circumvent existing safety constraints---such as removing step or cost limits---or inadvertently deletes parts of the existing scaffold, risking unintended system behavior. \\
\bottomrule
\end{tabular}
}
\vspace{10pt}
\caption{Error types of optimizer errors.}
\label{tab:error-taxonomy}
\end{table*}

\begin{table*}[h]
\centering
\resizebox{0.9\linewidth}{!}{%
\begin{tabular}{p{0.23\linewidth} p{0.16\linewidth} p{0.52\linewidth}}
\toprule
\textbf{Error Type} & \textbf{Domain} & \textbf{Example} \\
\midrule
Redundant Duplication & GAIA &
The optimizer adds a \texttt{validation\_solution\_txt} tool even though an almost identical \texttt{answer\_format\_guard} tool already exists. \\
\midrule
Hardcoding & $\tau^2$-Bench &
The optimizer hardcodes benchmark-specific values such as requiring an \texttt{excellent} speed-test result or refilling exactly \texttt{2GB} of mobile data. \\
\midrule
Task-specific Addition & GAIA &
The optimizer adds an award-lookup rule that assumes works released in year Y are awarded in year Y+1, making the scaffold rely on a brittle task-specific convention rather than a generally valid award-resolution strategy. \\
\midrule
Hallucination & Spider 2.0-Lite &
The optimizer writes a memory instructing the agent to use a \texttt{grouped\_average\_ranker} tool, but no such tool exists in the harness. \\
\midrule
Overengineering & $\tau^2$-Bench &
The optimizer converts a simple telecom troubleshooting decision rule into a dedicated \texttt{telecom\_next\_action} tool, adding unnecessary tool-use overhead. \\
\midrule
Direct Performance-degrading Update & SWE-Bench Verified&
The optimizer removes the existing ReAct-style interaction format and replaces it with a rigid single-command format, causing task success to drop sharply. \\
\midrule
Overgeneralized Heuristic & SWE-bench Verified&
The optimizer treats any helper-tool \texttt{SyntaxError} as likely Python-version incompatibility, even though syntax errors can arise from many unrelated causes. \\
\midrule
Safety Violation & SWE-bench Verified&
The optimizer removes the step and cost limits from the scaffold, weakening basic resource-control safeguards during task execution. \\
\bottomrule
\end{tabular}%
}
\caption{Representative cases for each error category.}
\label{tab:error-cases}
\end{table*}

\subsection{Analysis III}

To evaluate optimizers' awareness of the consequences of their actions (Table~\ref{tab:preliminary_binary}), we provide each optimizer with the current harness $h_t$ and the updated harness $h_{t+1}$, and prompt the same LLM brain of the optimizer to predict in natural language whether the update $h_t\rightarrow h_{t+1}$ improves or hinders the target agent's performance, i.e., whether $\Delta r > 0$ or $\Delta r \leq 0$.
That is, for updates produced by the GPT-5.2-based optimizer, we query GPT-5.2 for the prediction, and similarly for Claude Sonnet 4.6 and Qwen3.6-Plus.
The prompt used for this prediction task is shown in Figure~\ref{fig:evaluator_prompts}.
Using the harness updates collected in Appendix~\ref{app:preliminary_1_2}, we evaluate prediction performance.
For each domain, we test the optimizers on 11 pairs of consecutive harnesses.

\begin{figure}[t]
    \centering
    \includegraphics[width=1\linewidth]{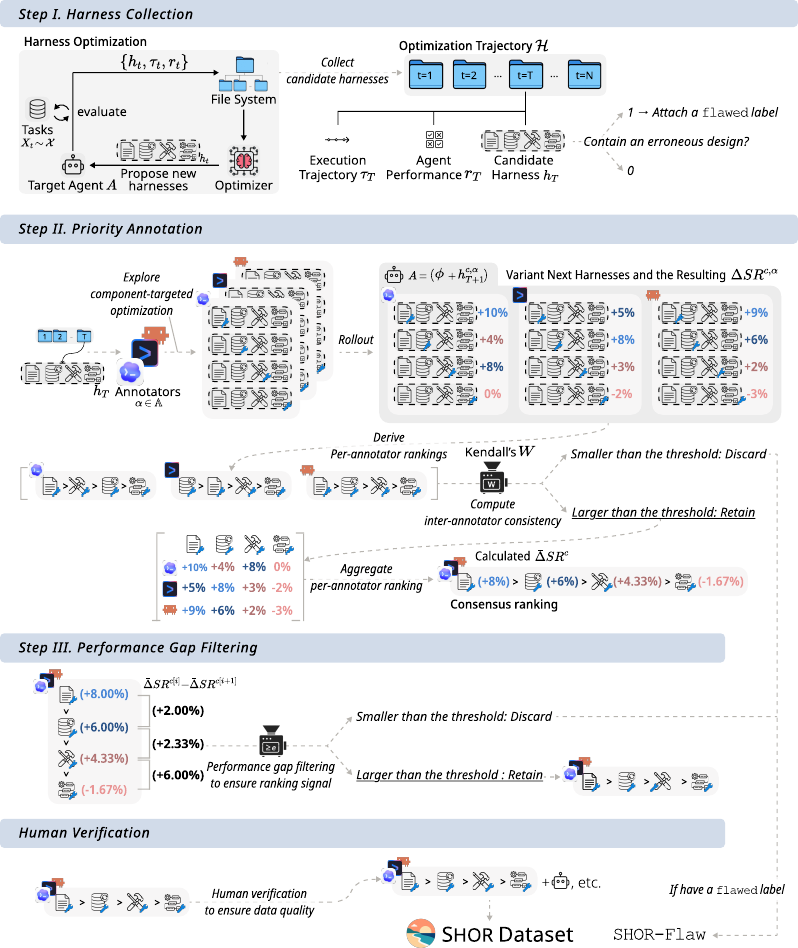}
    \vspace{15pt}
    \caption{Illustration of the dataset collecting process.}
    \label{fig:dataset_generation}
\end{figure}

\section{Details on \shor~\textsc{SHOR} Dataset}
\label{app:SHOR}
\paragraph{Illustration.} The illustration of \textsc{Shor}'s collecting process is present in Figure~\ref{fig:dataset_generation}.

\paragraph{Details on Collecting Candidate Harnesses for \textsc{Shor}.}
We adopt Meta-Harness to collect candidate harnesses $h_{\text{Cand}} = \{h^{(j)}_T\}_{j=1}^{J}$ following the procedure in~\citet{lee2026meta}.
At each iteration $t$, the optimizer $\textrm{OPT}$ observes the optimization history $\mathcal{H}_T = \{(h_i, \tau_i, r_i, S_i)\}_{i=1}^{T}$, stored as a filesystem, and proposes $k = 3 \sim 5$ new harness candidates $\{h_{t+1,1},h_{t+1,2},h_{t+1,3},\cdots\}$. 
Each candidate is first verified for successful execution (\ie, runs without errors), and only those that pass are further evaluated.
The candidates are then run on a valid set of 4 to 12 randomly sampled instances to collect $\tau_t$ and $r_t$, using gpt-4.1-mini as the target agent.
The optimization loop repeats for 20 iterations.
Final performance is separately measured on a held-out test set of 20 instances. 
Also, unlike the original Meta-Harness, which retains only harnesses on the Pareto frontier, we retain all intermediate harnesses.
As a result, we collected a total of J=\texttt{595} candidate harnesses across domains and time steps.
During the process, we also attach a \texttt{flawed} label to $h$ if it contains an erroneous design (determined based on domain information and skills, as in Section~\ref{sec:preliminaries} and Appendix~\ref{app:preliminary}).

\begin{figure}[ht!]
  \centering
  \includegraphics[width=0.9\textwidth]{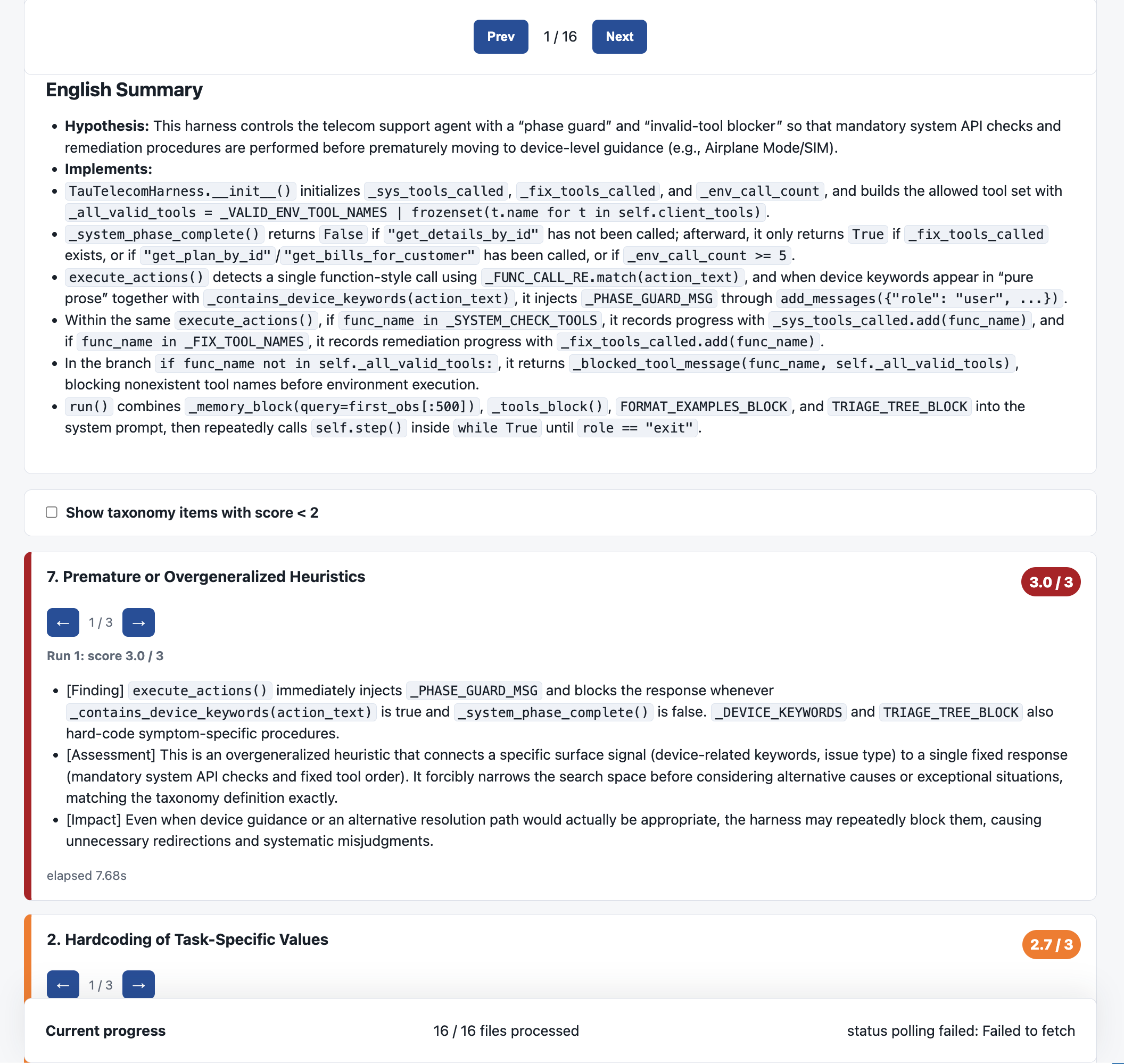}
  \caption{\texttt{Flaw} annotation tool interface.}
  \label{fig:flaw_annotation_tool}
\end{figure}

During the process, we also attach a \texttt{flawed} label to $h$ if it contains an erroneous design (determined based on domain information and skills, as in Section~\ref{sec:preliminaries} and Appendix~\ref{app:preliminary}).
To assign \texttt{flawed} labels reliably, we employ an LLM-assisted annotation pipeline.
Specifically, we use GPT-5.5 to score the extent to which each harness exhibits flaws with respect to every category in our taxonomy (Section~\ref{sec:preliminaries}).
To mitigate stochasticity, we generate three independent score sets per harness and average them.
The averaged scores are then provided to human annotators as a reference, who inspect the harness and assign the final \texttt{flawed} label corresponding to the most problematic component (Figure~\ref{fig:flaw_annotation_tool}).
This procedure combines the scalability of LLM-based assessment with human judgment, yielding labels that are both consistent and grounded in expert verification.

\subsection{Priority Annotation and Filtering.}
\paragraph{Agent Annotators.}
We adopt three SOTA coding agent as annotators for dataset collection~\textemdash~Codex (GPT-5.3-Codex); Claude Code (Claude-Sonnet-4.6); Gemini-CLI (Gemini-3-Pro)~\textemdash~as described in Step II, Section~\ref{sec:our_dataset}.
The prompt used to instruct each annotator is provided in Figure~\ref{fig:annotator_prompt_template}.

\paragraph{Computing Agent Success Rate for Ranking Signals.}
To acquire per-annotator ranking, we compute $\Delta\text{SR}^{c,\alpha}$ in Equation~\ref{eq:per_annotator_ranking} by running the target agent $A$ with each variant harness $h^{c,\alpha}_{T+1}$ on 20 task instances. 
We use Qwen-3.5-Flash as the LLM brain $\phi$ of $A$, applying a single rollout per instance.
The same set of task instances is shared across all annotators to ensure fair comparison.

\paragraph{Filtering via Kendall's W.}
To ensure label reliability, we compute Kendall's W among ${R^\alpha}_{\alpha \in \mathbb{A}}$ and discard $h_T$ with $W \leq \epsilon$, retaining only those with strong inter-annotator consistency.
We set $\epsilon = 0.500$, retaining 238 out of 595 instances after filtering.
Filtered instances with \texttt{flawed} label are added to \texttt{SHOR-Flaw}, which is a subset of \textsc{Shor} consisting of instances whose base harness contains erroneous designs.

\paragraph{Filtering via Performance Gap.}
We apply a final filtering step to ensure that only instances with strong ranking signals are collected for \textsc{Shor}. 
We require that adjacently ranked components in $R$ have a sufficiently distinct mean performance gap ($\overline{\Delta}\text{SR}^{c_{[i]}} - \overline{\Delta}\text{SR}^{c_{[i+1]}} > \delta$), and set $\delta = 0.005$. 182 out of 238 instances are remained after filtering.
Among the filtered-out instances, those with \texttt{flawed} label are also added to \texttt{SHOR-Flaw}.

\paragraph{Human Verification.}
We manually verify every above harness and its annotation quality before including it as a base harness $h$ in \textsc{Shor}. 
Finally, 182 harnesses are collected and 108 of which are attached with a \texttt{flawed} label.

\subsection{Dataset Statistics}
\label{app:shor_stat}

\begin{figure}[t]
  \begin{subfigure}[b]{0.46\textwidth}
    \centering
    \resizebox{\linewidth}{!}{%
    \begin{tabular}{lc}
      \toprule
      \textbf{Metric} & \textbf{Value} \\
      \midrule
      Total Instances & 182 \\
      \midrule
      Flawed Harnesses (\texttt{SHOR-Flaw}) & 122 \\
      \midrule
      Number of Domains & 4 \\
      \midrule
      Avg. Inter-annotator Consistency ($\omega$) & 0.7192 \\
      \bottomrule
    \end{tabular}%
    }
    \vspace{0.3cm}
    \caption{\textsc{Shor} Dataset Statistics}
    \label{tab:shor-stats}
  \end{subfigure}
  \hfill
  \begin{subfigure}[b]{0.46\textwidth}
    \centering
    \includegraphics[width=0.85\textwidth]{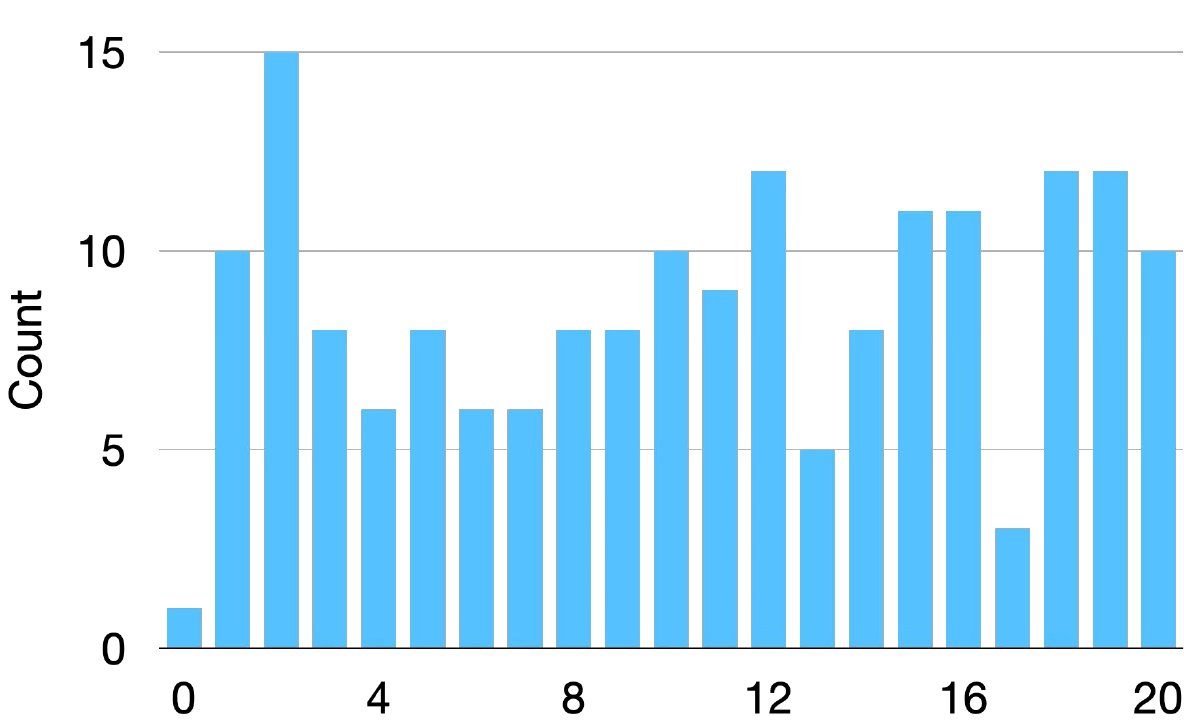}
    \caption{Number of instances per timestep}
    \label{fig:timestep_bar_graph}
  \end{subfigure}
  \caption{Overview of the \textsc{Shor} dataset, including key statistics (left) and the distribution of instances across timesteps (right).}
  \label{fig:combined}
\end{figure}

\begin{figure}[b]
  \centering
  \includegraphics[width=\textwidth]{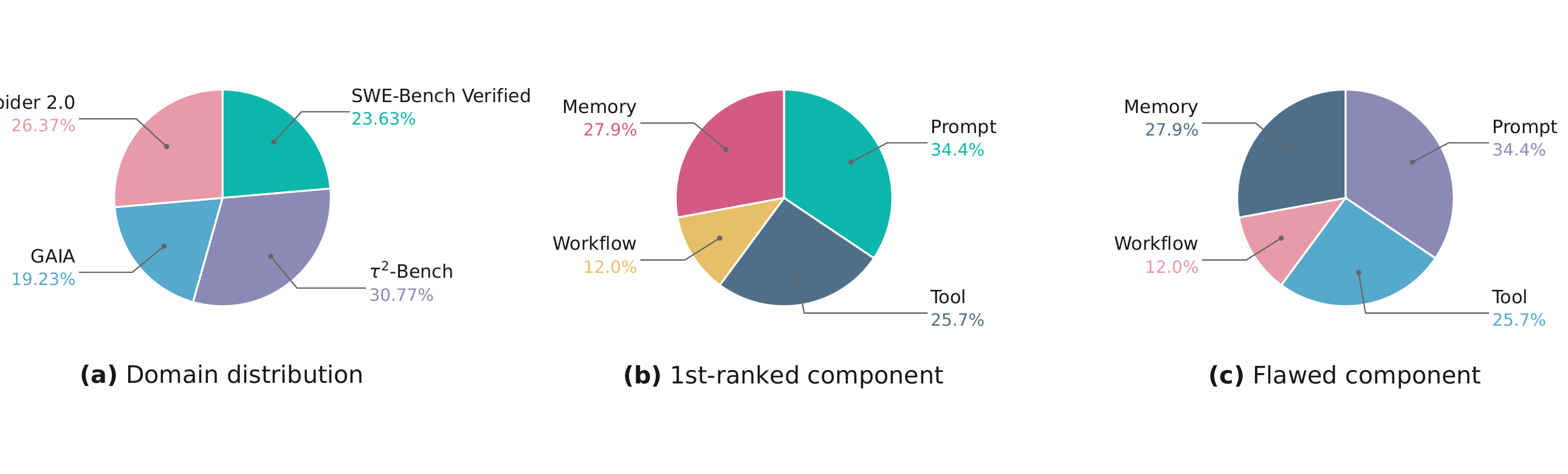}
  \vspace{-15pt}
  \caption{Distribution statistics of \textsc{Shor}.}
  
  \label{fig:distributions}
\end{figure}

Table~\ref{tab:shor-stats} shows an overview of \textsc{Shor}, including the total number of instances, number of flawed harnesses (\texttt{SHOR-Flaw}), number of domains, and average inter-annotator consistency.
Figure~\ref{fig:timestep_bar_graph} presents the number of instances per timestep. 
Figure~\ref{fig:distributions} (a) illustrates the domain distribution.
The distribution of the 1st-ranked components and flawed components are shown in Figures~\ref{fig:distributions} (c) and (b), respectively.

\paragraph{Auxiliary Field $I$.}
\label{app:aux}
The auxiliary field $I$ comprises three groups: metadata, a human-annotated quality label, and by-products from the annotation process.
\textbf{Metadata} records the contextual information of each instance, including the base harness performance $r^A(\tau_T, x_T)$, the domain and task identifier, and the hyperparameters used during annotation ($\epsilon, \delta$).
\textbf{Quality Label} records whether the base harness $h_T$ is \texttt{Flawed} or not, as determined by human annotators during the verification step.
\textbf{Artifacts} are the intermediate outputs generated throughout the annotation process (Section~\ref{ssec:annotating}). For \texttt{Flawed} data, \textbf{Artifacts} also include an explanation of their design errors. 
For each component $c \in \mathcal{C}$ and annotator $\alpha \in \mathcal{A}$, we store the component-targeted next harness $h^{c, \alpha}_{T+1}$, its execution trajectory $\tau^{c,\alpha}_{T+1}$, the resulting $\text{SR}^{c,\alpha}=r^A(\tau^{c,\alpha}_{T+1}, x_T)$, and the corresponding $\Delta\text{SR}^{c,\alpha} = r^A(\tau^{c,\alpha}_{T+1}, x_T) - r^A(\tau_T, x_T)$.
From these, we further store the per-annotator ranking $R_\alpha$ derived by sorting $\mathcal{C}$ in descending order of $\Delta\text{SR}^{c,\alpha}$, the inter-annotator consistency score $\omega$ (Kendall's $W$) computed across $\{R_\alpha\}_{\alpha\in\mathcal{A}}$, and the adjacent performance gaps $\bar{\Delta}\text{SR}^{c_{[i]}} - 
\bar{\Delta}\text{SR}^{c_{[i+1]}}$ 
for all $i \in [1, N-1]$.

\section{Details on Harness Optimizers and Agent Applied in This Study}
\label{app:optimizers}
\subsection{Optimizers under Evaluation}

\paragraph{Claude Code.}
Claude Code~\citep{claude2026claudecode} is an agentic coding system that operates directly in the terminal, designed to autonomously complete multi-step software development tasks. 
It serves as a harness around a Claude language model, providing tools, context management, and an execution environment that transforms the underlying LLM into a capable coding agent. 
The system follows an iterative agent loop: reading project files, analyzing context, proposing an action, executing approved tools, and evaluating results.
To maintain persistent context across sessions, Claude Code supports \texttt{CLAUDE.md} — a repository-level instruction file that specifies project conventions, build commands, and workflow preferences, automatically loaded at the start of every session.
Unlike code completion tools, Claude Code operates at the project level: reading full codebases, planning across multiple files, executing changes, running tests, and iterating on failures, making it well-suited as an autonomous optimizer agent in automated evaluation pipelines.

\paragraph{Meta-Harness.}
Meta-Harness~\citep{lee2026meta} is built upon Claude Code. It automatically optimizes harness code for LLM agents, where a harness is a single-file Python program that wraps a model and determines how the model is invoked, including prompt construction, retrieval, memory updates, and orchestration logic.
The framework operates as an iterative search loop in which a coding agent, given access to a filesystem accumulating the history of prior candidates' source code, evaluation scores, and raw execution traces, freely inspects this history, diagnoses failure modes, and proposes new harness candidates at each iteration.
The key design principle is to expose full history without compression, enabling the proposer to trace downstream failures back to specific earlier design decisions rather than reacting to lossy feedback signals. 
The optimization loop uses a Pareto-frontier-based algorithm to search over executable agent designs, replacing human craftsmanship.

\paragraph{mini-swe-agent.}
mini-swe-agent~\citep{mini-swe-agent} is a minimalist LLM-based software engineering agent implemented in approximately 100 lines of Python. 
Unlike its predecessor, SWE-agent~\citep{yang2024swe}, which relied heavily on specialized tools and custom agent-computer interfaces, mini-swe-agent operates solely via bash commands, without any additional tooling or tool-calling interface, making it compatible with virtually any LLM brain.
The agent maintains a fully linear interaction history in which each step appends to the message list.
Every action is executed independently via \texttt{subprocess.run}. 
Despite its minimal design, mini-swe-agent achieves more than 74\% on SWE-bench Verified, demonstrating that strong SWE performance can be attained without complex scaffolding. 
Its transparency and minimal dependencies make it a useful baseline for research on LLM agents.

\paragraph{ReCreate.}
ReCreate~\citep{hao2026recreate} is built upon mini-swe-agent. It's an experience-driven framework that automatically creates domain agents by optimizing agent harnesses from interaction experience, rather than relying solely on performance metrics. 
Specifically, ReCreate iteratively inspects the target agent's interaction trajectories, execution logs, and environment states to propose harness updates across four editable components: system prompt, instance prompt, tools, and memory. 
ReCreate introduces three key components: (i) on-demand experience retrieval mechanism that allows ReCreate to actively navigate and inspect critical events rather than processing the full experience at once; (ii) a reasoning-creating synergy pipeline where a creation router grounds every harness update in specific execution evidence; and (iii) a hierarchical local-to-domain update mechanism that buffers instance-level proposals and aggregates them into reusable domain patterns, preventing overfitting to individual tasks.
Together, these three components enable ReCreate to automatically create domain agents from minimal seed harnesses.

\paragraph{OpenHands-CLI.}
OpenHands-CLI is a lightweight terminal interface built on top of the OpenHands~\citep{wang2024openhands}, an open-source platform for developing generalist AI software agents that interact with the world as a human developer would, such as writing code, executing commands, and browsing the web. 
The platform provides a sandboxed Docker runtime in which all actions are executed and returned as observations through an event stream architecture, with each agent implementing a \texttt{step()} function that maps the current state~\textemdash~a chronological history of past actions and observations~\textemdash~to a new action. 
OpenHands-CLI uses CodeActAgent~\citep{wang2024executable} as its default agent, which consolidates agent actions into a unified code action space — executing bash commands, Python code, and browser interactions — rather than relying on JSON-based tool calls. We follow this default setting.

\paragraph{Codex.}
OpenAI Codex~\citep{openai2025codex} is a SWE agent built on OpenAI models, designed to automate complex, multi-step coding tasks. 
Codex also follows an iterative agent loop, cycling through model inference, tool execution, and result evaluation until the task is complete.
Codex can write features, fix bugs, answer codebase questions, and propose pull requests. 
Codex supports repository-level instruction files (\texttt{AGENTS.md}) that specify codebase navigation, testing commands, and project conventions, enabling structured task delegation without manual prompting. 
This design positions Codex as an agentic system well-suited for autonomous, end-to-end execution of real-world SWE workflows.

\paragraph{Gemini CLI.}
The Gemini Command-Line Interface (CLI)~\citep{geminicli2026geminicli} is an open-source agentic coding assistant developed by Google, designed to operate directly within a terminal environment.
Gemini CLI also follows an iterative agent loop, cycling through reasoning, tool invocation, and result evaluation to complete complex software engineering objectives such as bug fixing, feature implementation, and test generation.
Built-in tools include file operations, shell command execution, web fetching, and Google Search grounding, enabling the agent to interact with both local environments and external resources.
For broad tasks, Gemini CLI acts as a strategic orchestrator, delegating sub-tasks to specialized subagents that operate in isolation with their own tool sets and context windows, consolidating results back to the primary agent.

\subsection{Agent Used as Annotators}
Within the optimizers introduced above, three of them are also adopted as annotators for dataset collection~\textemdash~Codex (GPT-5.3-Codex); Claude Code (Claude-Sonnet-4.6); Gemini-CLI (Gemini-3-Pro)~\textemdash~as described in Step II, Section~\ref{sec:our_dataset}.

\subsection{Large Language Models}
We list all LLMs used across different phases of dataset construction and evaluation in Table~\ref{tab:llms}.

\begin{table}[h]
\centering
\large
\resizebox{\textwidth}{!}{
\begin{tabular}{lccccc}
\toprule
\textbf{Model} & \textbf{Preliminary Analysis} & \textbf{Harness Collection} & \textbf{Priority Annotation} & \textbf{\textsc{Shor} Evaluation} & \textbf{Harness Optimization} \\
\midrule
Claude Haiku 4.5~\cite{anthropic2025haiku45}       & & & & \checkmark & \checkmark \\
Claude Sonnet 4.6~\cite{anthropic2026sonnet46}      & \checkmark & \checkmark & \checkmark & \checkmark & \checkmark \\
GPT-4.1-Mini~\cite{openai2025gpt41mini}             & & $\checkmark^\dagger$ & & & \\
GPT-5-Mini~\cite{openai2025gpt5}                    & $\checkmark^\dagger$ & & & & $\checkmark^\dagger$ \\
GPT-5.2~\cite{openai2025gpt52}                      & \checkmark & & & \checkmark & \checkmark \\
GPT-5.3-Codex~\cite{openai2026gpt53codex}           & & & \checkmark & & \\
GPT-5.5~\cite{openai2026gpt55}                      & & & & \checkmark & \checkmark \\
Gemini 3 Flash~\cite{deepmind2025gemini3flash}       & & & & \checkmark & \checkmark \\
Gemini 3 Pro~\cite{deepmind2025gemini3pro}           & & & \checkmark & \checkmark & \checkmark \\
DeepSeek-V4-Pro~\cite{deepseekAI2026v4}             & & & & \checkmark & \checkmark \\
Qwen3.5-Flash~\cite{qwenteam2026qwen35}             & & & $\checkmark^\dagger$ & & \\
Qwen3.5-397B-A17B~\cite{qwenteam2026qwen35}         & & & & \checkmark & \checkmark \\
Qwen3.6-Plus~\cite{qwenteam2026qwen36plus}          & \checkmark & & & & \\
Kimi K2.6~\cite{moonshot2026kimik26}                & & & & \checkmark & \checkmark \\
GLM-5.1~\cite{glm5team2026glm5vibecodingagentic}    & & & & \checkmark & \checkmark \\
\bottomrule
\end{tabular}
}
\vspace{2pt}
\caption{LLMs used in each phases. $\dagger$ denotes models used as $\phi$ of the target agent.}
\label{tab:llms}
\end{table}

\subsection{Codes and Prompts}
\label{app:optim_code_prompt}

\paragraph{Codes.} 
\url{https://github.com/k59118/Harness_Optimizer_Evaluation}.

\paragraph{Prompts.}

\begin{itemize}
    
    \item The \textit{harness update prompt} for ReCreate-Agent, used in Section~\ref{sec:preliminaries} is in Appendix~\ref{app:recreate_prompt}.
    
    \item The \textit{update action awareness prompt} for ReCreate-Agent used in Section~\ref{sec:preliminaries} is in Appendix~\ref{fig:evaluator_prompts}.
    
    \item The \textit{harness collection prompt} for Meta-Harness used in Section~\ref{ssec:collecting_base_harness} is in Appendix~\ref{app:component_targeted_optimization_prompt}.
    
    \item The \textit{component-targeted optimization prompt} for annotators $\alpha\in\mathbb{A}$ (Claude Code, Codex, Gemini-CLI) used in Section~\ref{ssec:annotating} is in Appendix~\ref{app:component_targeted_optimization_prompt}.
    
    \item The \textit{harness optimizer prompt}  used in Section~\ref{sec:evaluation_settings} is in Appendix~\ref{app:harness_optimizater_prompt}.

    \item The \textit{\textsc{Shor} evaluation prompt} used in Section~\ref{sec:evaluation_settings} is in Figure~\ref{app:priority_ranking_prompt}.

\end{itemize}

% = = ====================================================

\section{Details on Domains and Datasets}
\label{app:datasets}

\subsection{In-domain Environments}
\textbf{Spider 2.0}. Proposed by \citet{lei2024spider}, it is a collection of 632 text-to-SQL workflow problems derived from real-world enterprise-level cases. Databases in Spider 2.0 are sourced from real industrial applications such as Google Analytics and feature massive schema items (an average of 812 columns) with unique structures (\eg, nested columns), along with terabyte-scale data volumes. These databases range from local databases (\eg, SQLite and DuckDB) to cloud data warehouses (\eg, BigQuery and Snowflake). Problems in Spider 2.0 require agents to understand and search across database metadata, dialect documentation, and project-level codebases, challenging agents to reason over complex SQL workflow environments and extremely long contexts.

\noindent $\tau^{2}$\textbf{-Bench.} Present by~\citet{barres2025tau}, it is an extension of the existing $\tau$-bench~\citep{yao2024tau}, and its core innovation is the dual-control environment. Prior benchmarks are all ``single-control'', that is, only the AI agent can use tools to act in the world, while the simulated user is a passive information provider. However, real-world customer support scenarios such as telecom troubleshooting require users to actively participate in the task (restart a phone, toggle airplane mode, reseat a SIM card). $\tau^{2}$-Bench addresses this by giving the simulated user their set of tools to act on the shared environment. This dataset covers the retail, airline, and telecom domains, and we apply the telecom portion.

\noindent \textbf{{GAIA.}} 
Proposed by~\citet{mialon2023gaia}, it is a benchmark for evaluating General AI Assistants on real-world, verifiable questions. 
It consists of 466 unique questions, carefully annotated by humans across diverse assistant use cases.
Unlike prior benchmarks targeting tasks that are increasingly difficult for humans, GAIA focuses on conceptually simple yet practically challenging tasks that require accurate execution of complex action sequences. 
Specifically, it evaluates agents against 4 core capabilities: Reasoning (multi-step inference to derive a correct answer), Web Browsing (searching and navigating the open web), Multi-modality (handling inputs beyond text such as images, audio, and video), and Tool Use (proficiency with diverse tools including code interpreters and file readers). 
Questions are categorized into three difficulty levels based on the number of steps and tools required, from Level 1 (minimal tools, up to 5 steps) to Level 3 (arbitrary sequences of actions with unrestricted tool access).

\noindent \textbf{SWE-bench Verified.} Collected by~\citet{openai2024swe}, it is a human-verified dataset consisting of 500 software engineering problems derived from SWE-bench~\citep{jimenez2023swe}. The problems are collected from real GitHub issues and corresponding pull requests across 12 Python repositories. Issues in SWE-bench often require agents to understand and coordinate changes across functions, classes, and files at the same time, necessitating complex reasoning that goes beyond traditional code generation tasks.

\subsection{Out-of-domain Environments}

\textbf{GPQA.}
Proposed by Rein et al.~\citep{rein2024gpqa}, it is a dataset of 448 multiple-choice questions written by domain experts in biology, physics, and chemistry, targeting graduate-level difficulty. 
The benchmark's central design principle is being "Google-proof": the questions are crafted to resist lookup-based shortcuts, demanding genuine scientific understanding rather than surface-level recall. 
Even PhD-level non-expert validators, given unrestricted internet access, spent an average of 37 minutes per question yet achieved only 34\% accuracy, while in-domain experts reached 65\%. 
This sharp performance gap between experts and non-experts makes GPQA particularly valuable for scalable oversight research, probing whether humans can reliably verify AI outputs in domains where AI capability may approach or exceed human expertise. 
The benchmark also includes a curated Diamond subset of 198 questions with the highest inter-expert agreement, which has become the de facto standard for frontier model evaluation.

\textbf{AppWorld.}
Presented by Trivedi et al.~\citep{trivedi2024appworld}, it consists of a high-fidelity execution environment of 9 day-to-day apps operable via 457 APIs, paired with a benchmark of 750 autonomous agent tasks. 
While prior tool-use benchmarks were limited to simple linear sequences of API calls, AppWorld requires agents to generate rich code with complex control flow through iterative interaction with the environment — for instance, managing a household's grocery orders across messaging, notes, and shopping apps simultaneously. 
Evaluation is conducted via state-based unit tests that verify not only task completion but also the absence of unintended side effects, \ie, collateral damage. 
The benchmark includes both a normal and a challenge test split, with even the strongest models at the time of release solving only around half of the normal tasks and roughly 30\% of the challenge tasks.

\subsection{Splits}
\paragraph{Correlation between priority ranking and full harness optimization.} 
When evaluating optimizer performance in actual multi-iteration harness optimization (Figure~\ref{fig:in_correlation}), we take 45 data instances from each domain/datasets as the test set to compute target agents' performance. 
These test sets do not overlap with tasks we use to build \textsc{Shor}.

\paragraph{Correlation between priority ranking and error recovery.}
To test the optimizer's ability to resolve errors (Table~\ref{tab:optimizer-harness-resolve}), we additionally collect a test set of \texttt{Flawed} harnesses. 
The task instances used to derive these harnesses do not overlap with the tasks we use to collect \textsc{Shor}.

\section{Details on Main Experiments}
\label{app:main_exp}

\subsection{\textsc{Shor} evaluation}
\label{app:shor_eval}

For each instance $d$ of \textsc{Shor} (see Section~\ref{ssec:annotating}), the optimizer is prompted to produce a natural-language output for each component ranking, with access to the full optimization trajectory $\mathcal{H}$ stored in the file system. 
The optimizer can inspect the trajectory as needed.
We compare the optimizer's outputs against the priority labels provided in \textsc{Shor} using four metrics:
Top-1 accuracy, which measures whether the optimizer correctly identifies the highest-priority component $c_{[1]}$; MRR, which computes the reciprocal rank of the first correctly ranked component; NDCG, which evaluates the quality of the full ranking by accounting for the position of each component; and Kendall's $\tau$, which measures the ordinal correlation between the predicted and ground-truth priority rankings.

\begin{figure}[t!]
  \centering
  \includegraphics[width=\textwidth]{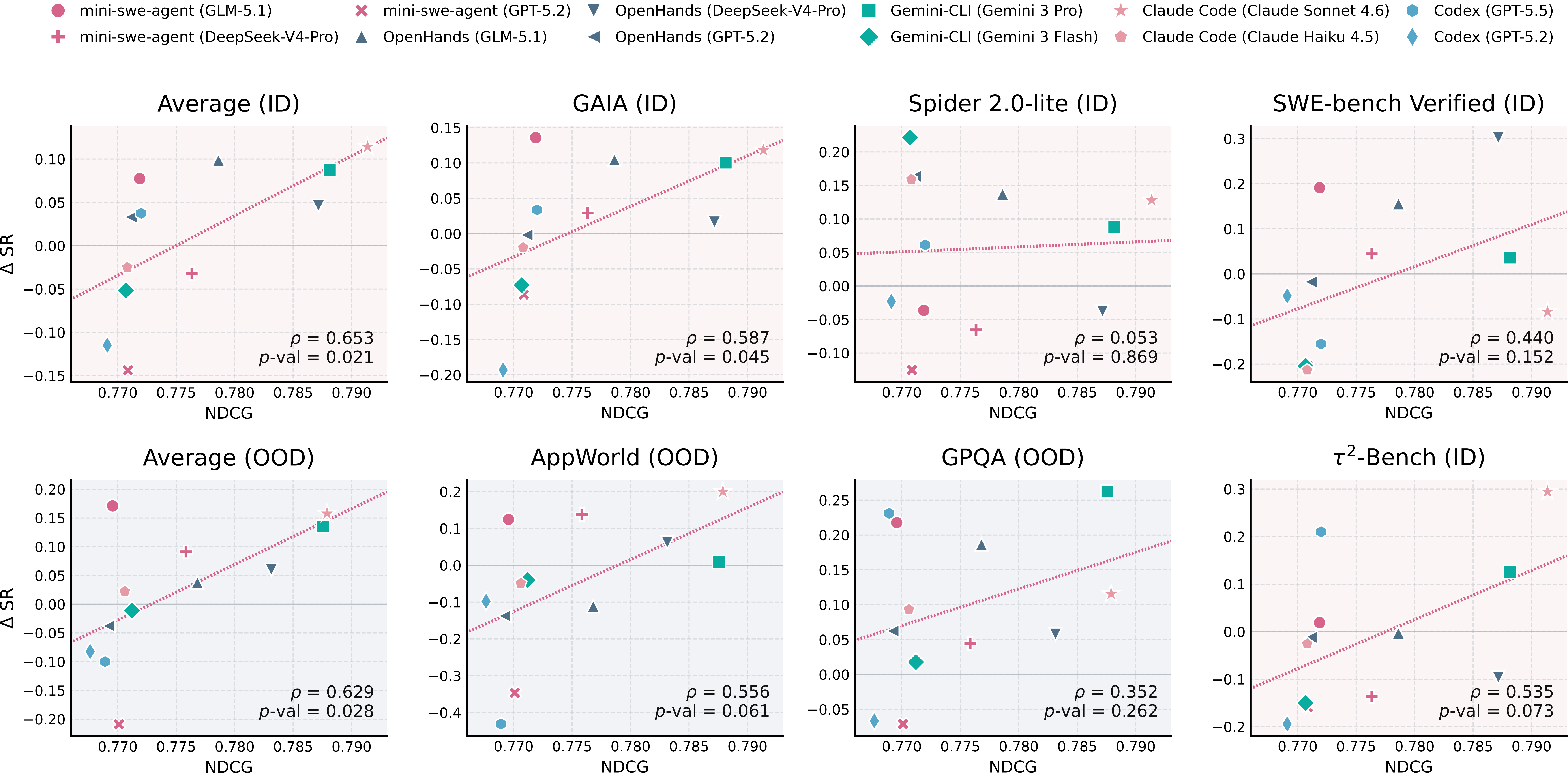}
  \vspace{2.5pt}
  \caption{Correlation between priority rankings and the optimizer’s ability to improve target agents’ SR in harness optimization, evaluated using the NDCG metric.}
  \label{fig:in_correlation_ndcg}
\end{figure}

\subsection{Harness Optimization}

We evaluate the same optimizers used in Appendix~\ref{app:shor_eval} on harness optimization across both in-domain and out-of-domain settings. 
For each domain, we randomly sample 5 base harnesses and use each as the initial harness of the target agent, where base harnesses are randomly sampled from the harnesses filtered in \textsc{Shor} for the in-domain setting, and collected using Meta-Harness following the same procedure as \textsc{Shor} construction for the out-of-domain setting.
During optimization, the target agent is evaluated on a validation set of 4 instances randomly sampled per domain, and the optimizer receives the current harness and the resulting trajectories as input.
The optimizer then optimizes the harness for 10 iterations, after which we compute $\Delta$SR as the difference in performance relative to the base harness on a held-out test set of 45 instances. 
The target agent is mini-swe-agent (harness) paired with gpt-5-mini (LLM brain). 

\subsection{Efficiency Analysis}
\label{app:cost_and_time}
We measure the efficiency of each optimizer along two axes: cost and time.

\paragraph{Cost Estimation.} We estimate cost based on token usage. 
For harness optimization, we report the average cost per optimizer when running 10 iterations across 4 domains and 5 base harnesses each. 
For \textsc{Shor} evaluation, we report the average cost per optimizer over all 140 instances. 
Note that while the actual number of evaluation instances used in harness optimization is 45, we normalize to 140 instances, matching the total number of \textsc{Shor} instances, to ensure a fair comparison.

\paragraph{Time Measurement.} 
We report wall-clock time for each optimizer. 
For harness optimization, we measure the total wall time from the start of optimization to the completion of the final harness, assuming sequential execution across 4 domains and 5 base harnesses each for 10 iterations. 
For \textsc{Shor} evaluation, we report the average wall time over all 140 instances. 
As with cost, we normalize the harness optimization time to 140 evaluation instances for a fair comparison.

\begin{figure}[t!]
  \begin{tcolorbox}[colback=gray!10, colframe=gray!50, title=Example $S_i$ (GAIA; Step 2)]
  \small{
    \textbf{GAIA agent}: refined prompt with keyword memory + SmartVerifyAgent.
    \newline
    \newline
    - PRIMARY\_AXIS: memory
    
    - SECONDARY\_AXES: prompt
    \newline
    \newline
    \textbf{Hypothesis}: \textbf{Replacing recency-based memory with keyword-overlap (Jaccard) retrieval} will surface more topically relevant past Q/A examples, preventing wrong-answer contamination.
    Combining this with an improved \texttt{VerifyLoopAgent} that strips inline Wikipedia boilerplate and detects format pollution will catch empty and misformatted answers before submission, pushing accuracy above 50\%.
    \newline
    \newline
    \textbf{Changes vs gaia\_refined\_prompt}:
    \newline
    - \texttt{RecentMemoryRetriever} replaced with \texttt{KeywordMemoryRetriever} (Jaccard unigram overlap scoring)
    \newline
    - \texttt{VerifyLoopAgent} enhanced with:
        \newline
        \phantom{--} * \texttt{\_strip\_inline\_boilerplate()} to remove wiki markup (References, See also, etc.)
        \newline
        \phantom{--} * Broader non-answer phrase detection
        \newline
        \phantom{--} * \texttt{\_read\_and\_clean\_solution()} to check for common format failures before detection
    \newline
    - \texttt{\_store\_memory\_result}: skips non-Submitted and empty-answer records to prevent memory contamination from failed runs.
    }
  \end{tcolorbox}
  \caption{An example optimizer summary $S_i$ generated at step $i=2$ of a GAIA trajectory, illustrating how the optimizer diagnoses failure modes and proposes targeted harness updates.}
  \label{fig:example_summary}
\end{figure}

\subsection{Priority Ranking as an Actionable Insight}

In the experiment, we use \texttt{SHOR-Flaw} as the source of flawed harnesses.
The oracle information given to the optimizer in the second setting consists of the target flawed code segment, the human-annotated description of the problem, and the corresponding metadata from the instance of \texttt{SHOR-Flaw}.
After the resolution, human annotators inspect both the original flawed harness and the resulting harnesses to determine whether each flaw has been resolved. 
Resolving performance is reported as the proportion of harnesses judged to be successfully fixed.

\section{Other Details}
\label{app:other_details}

\paragraph{Example Summary $S_i$.}

Below is an example summary $S_i$ written by the optimizer at step $i$, drawn from a GAIA trajectory.
$S_i$ captures the optimizer's analysis of the target agent's behavior and informs subsequent harness updates.

As shown in Figure~\ref{fig:example_summary}, it demonstrates how $S_i$ provides structured insights into agent failure patterns and optimization priorities, directly informing the next harness iteration. Note that $S_1 = \emptyset$ since no prior trajectory exists at the initial step.

\section{Additional Results}
\label{app:additional_results}
\subsection{Correlation Analysis}
We use optimizers' performance in priority ranking, reported via NDCG, to draw the same correlation plot shown Figure~\ref{fig:in_correlation} and present it in Figure~\ref{fig:in_correlation_ndcg}.

\subsection{Generalizability of the correlation between priority ranking and agent improvement}
Following the same analysis in Figure~\ref{fig:in_correlation}, we change the target agent to mini-swe-agent (Qwen-3.5-flash) and present the correlation result Table~\ref{tab:change_brain}. Due to budget limits, we present only 2 in-domain and 1 OOD task.

\begin{table}[t]
\centering
\small
\setlength{\tabcolsep}{8pt}
\renewcommand{\arraystretch}{1.15}
\begin{tabular}{l cc cc}
\toprule
\textbf{Brains:} 
& \multicolumn{2}{c}{\textbf{GPT-5-mini (Current)}} 
& \multicolumn{2}{c}{\textbf{Qwen-3.5-flash (New)}} \\
\cmidrule(lr){2-3}
\cmidrule(lr){4-5}
& $\rho$ & \textit{p}-value
& $\rho$ & \textit{p}-value \\
\midrule
\textbf{SWE-V}
& 0.649 & 0.022
& 0.648 & 0.023 \\

\textbf{$\tau^2$-Bench}
& 0.342 & 0.277
& 0.399 & 0.120 \\

\textbf{Appworld}
& 0.600 & 0.039
& 0.556 & 0.061 \\
\bottomrule
\end{tabular}
\vspace{15pt}
\caption{Generalization to a different LLM brain in the target agent. We compare the result reported earlier and a new result using Qwen-3.5-flash as the LLM brain of the target agent whose harness is being optimized by the harness optimizer. $\rho$ indicates the Pearson correlation between optimizers' performance in priority ranking (ACC@1) and the target agent's $\Delta$SR in the task domain after the optimization loop.}
\label{tab:change_brain}
\end{table}

\begin{table}[t]
\centering
\small
\setlength{\tabcolsep}{8pt}
\renewcommand{\arraystretch}{1.15}
\begin{tabular}{l cc cc}
\toprule
& \multicolumn{2}{c}{\textbf{Current 12 Optimizers}}
& \multicolumn{2}{c}{\textbf{10 Current + 2 New Optimizers}} \\
\cmidrule(lr){2-3}
\cmidrule(lr){4-5}
\textbf{Domains}
& \textbf{Pearson $\rho$} & \textbf{\textit{p}-value}
& \textbf{Pearson $\rho$} & \textbf{\textit{p}-value} \\
\midrule
SWE-V
& 0.649 & 0.022
& 0.620 & 0.032 \\

$\tau^2$-Bench
& 0.342 & 0.277
& 0.314 & 0.320 \\

Appworld
& 0.600 & 0.039
& 0.476 & 0.139 \\
\bottomrule
\end{tabular}
\vspace{15pt}
\caption{We compare the result reported earlier and a new result where we replace two optimizers used in the data curation pipeline with Mini-swe-agent (gpt-5.6-luna) and OpenHands-CLI
(Gemini-3.6-flash). $\rho$ indicates Pearson correlation between optimizers' performance in priority ranking (ACC@1) and the target agent's $\Delta$SR in the task domain after the optimization loop.}
\label{tab:swap_optimizer}
\end{table}

\begin{table*}[t]
\centering
\vspace{5em}
\small
\resizebox{1\linewidth}{!}{
\setlength{\tabcolsep}{3.5pt}
\begin{tabular}{l|l|ccc|ccc|ccc|ccc|ccc}
\toprule
\multicolumn{2}{c|}{Harness Optimizer ($\textrm{OPT}$)}
& \multicolumn{3}{c|}{GAIA}
& \multicolumn{3}{c|}{Spider 2.0-lite}
& \multicolumn{3}{c|}{SWE-V}
& \multicolumn{3}{c|}{$\tau^2$-Bench}
& \multicolumn{3}{c}{\textbf{Average}} \\
\cmidrule(lr){3-5} 
\cmidrule(lr){6-8} 
\cmidrule(lr){9-11} 
\cmidrule(lr){12-14} 
\cmidrule(lr){15-17}

$\textrm{OPT}$'s Harness & $\textrm{OPT}$'s Base LLMs
& \XSolidBrush & \Checkmark & \FiveStar
& \XSolidBrush & \Checkmark & \FiveStar
& \XSolidBrush & \Checkmark & \FiveStar
& \XSolidBrush & \Checkmark & \FiveStar
& \XSolidBrush & \Checkmark & \FiveStar \\
\midrule

Gemini-CLI
& Gemini 3 Pro
& \cellcolor[HTML]{D2D8E8}{0.314}
& \cellcolor[HTML]{ABB6D6}{0.657}
& \cellcolor[HTML]{7F8EBF}{0.857}
& \cellcolor[HTML]{D2D8E8}{0.346}
& \cellcolor[HTML]{ABB6D6}{0.704}
& \cellcolor[HTML]{7F8EBF}{0.926}
& \cellcolor[HTML]{BFC8DF}{0.400}
& \cellcolor[HTML]{BFC8DF}{0.500}
& \cellcolor[HTML]{ABB6D6}{0.781}
& \cellcolor[HTML]{D2D8E8}{0.346}
& \cellcolor[HTML]{7F8EBF}{0.815}
& \cellcolor[HTML]{7F8EBF}{0.815}
& \cellcolor[HTML]{D2D8E8}{0.350}
& \cellcolor[HTML]{ABB6D6}{0.664}
& \cellcolor[HTML]{7F8EBF}{0.843} \\
\midrule

Claude Code
& Sonnet 4.6
& \cellcolor[HTML]{E1E5F0}{0.171}
& \cellcolor[HTML]{D2D8E8}{0.343}
& \cellcolor[HTML]{7F8EBF}{0.886}
& \cellcolor[HTML]{D2D8E8}{0.222}
& \cellcolor[HTML]{D2D8E8}{0.333}
& \cellcolor[HTML]{7F8EBF}{1.000}
& \cellcolor[HTML]{E1E5F0}{0.100}
& \cellcolor[HTML]{BFC8DF}{0.533}
& \cellcolor[HTML]{ABB6D6}{0.719}
& \cellcolor[HTML]{E1E5F0}{0.185}
& \cellcolor[HTML]{BFC8DF}{0.481}
& \cellcolor[HTML]{ABB6D6}{0.778}
& \cellcolor[HTML]{E1E5F0}{0.168}
& \cellcolor[HTML]{BFC8DF}{0.420}
& \cellcolor[HTML]{7F8EBF}{0.843} \\
\midrule

Codex
& GPT-5.5
& \cellcolor[HTML]{D2D8E8}{0.286}
& \cellcolor[HTML]{BFC8DF}{0.486}
& \cellcolor[HTML]{7F8EBF}{0.943}
& \cellcolor[HTML]{E1E5F0}{0.111}
& \cellcolor[HTML]{D2D8E8}{0.370}
& \cellcolor[HTML]{7F8EBF}{0.963}
& \cellcolor[HTML]{D2D8E8}{0.267}
& \cellcolor[HTML]{BFC8DF}{0.467}
& \cellcolor[HTML]{7F8EBF}{0.906}
& \cellcolor[HTML]{D2D8E8}{0.296}
& \cellcolor[HTML]{D2D8E8}{0.296}
& \cellcolor[HTML]{7F8EBF}{1.000}
& \cellcolor[HTML]{D2D8E8}{0.244}
& \cellcolor[HTML]{BFC8DF}{0.412}
& \cellcolor[HTML]{7F8EBF}{0.950} \\
\bottomrule
\end{tabular}
}
\caption{Human evaluation of optimizer performance (resolved rate) to resolve flawed harnesses that contain errors (\ie, the \texttt{SHOR-Flaw} dataset). \XSolidBrush and \Checkmark denote whether the location (\ie, which component) of serious flaws is provided. \FiveStar~ indicates that optimizers further have access to the ground-truth explanation of the flaws.}
\vspace{-0.5em}
\label{tab:optimizer-harness-resolve-full-oracle}
\end{table*}

To address concerns about annotator-optimizer overlaps, we replace Claude Code (Claude-Sonnet-4.6) and Gemini-CLI (Gemini-3-Pro) in Figure~\ref{fig:in_correlation} with Mini-swe-agent (gpt-5.6-luna) and OpenHands-CLI (Gemini-3.6-flash), and re-run the correlation experiments. 
Figure~\ref{tab:swap_optimizer} shows the correlation plot (due to budget limits, we present only 2 in-domain and 1 OOD task).

\subsection{Consistency between the Ranking Labels and Human Annotation}
The agreement on top-component between humans and SHOR labels is 57.8\% (avg. of 3 people), clearly above random (25\%) yet far from perfect. This gap is expected, as SHOR derives labels from actual $\Delta$SR while humans only rely on given harnesses/trajectories. This justifies our efforts to reduce human efforts in evaluating optimizers.

\subsection{Generalizability of the Data Curation Pipeline}
Using 32 [base harness, four variants] sets in SHOR, we re-ran our annotation pipeline (Qwen as the target-agent brains) to label the ranking of harness components. The consistency between these new rankings and the original ones in SHOR is:
\begin{itemize}
    \item Agreement@1 = 47.1\%: The percentage of new top-1 component that matches with the top-1 component.
\item Agreement@2 = 68.8\%: The percentage of new top-1 component was ranked as top-1 or -2 in the original annotation.
\end{itemize}
The result suggests the generalizability of our annotation process.

\subsection{Extended Result of Harness Optimizers' Performance in Error Recovery}
In Table~\ref{tab:optimizer-harness-resolve}, we present harness optimizers' performance in resolving erroneous harnesses when given the ground-truth information on which harness component contains an error. We hereby additionally present the results of a setting where the harness optimizers further receive the ground-truth explanations of the occurring errors in Table~\ref{tab:optimizer-harness-resolve-full-oracle}.

\section{Computing Resources}
\label{app:computing_resources}

Our experiments primarily utilized commercial API services rather than local computing resources. 
We accessed GPT-5.2 and GPT-5.5 through OpenAI's API, Claude Sonnet 4.6 and Claude Haiku 4.5 through Anthropic's API, and Gemini 3 Pro and Gemini 3 Flash through Google's API. 
For open-source models including Qwen, DeepSeek and GLM, we served them locally using 5 NVIDIA A100 GPUs. 
We have confirmed that all artifacts used in this paper are available for non-commercial scientific use.

\newpage
\section{Prompts}
\label{app:prompts}
\subsection{Prompts Used in Preliminary Analysis}
\label{app:recreate_prompt}

\begin{tcolorbox}[colback=gray!10, colframe=gray!50, title=System Prompt]
{\small
You are a \textbf{ReCreate-Agent} --- an agent that creates and evolves other AI agents by editing their scaffolds (prompts, workflows, tools, and memories).\\
Your mission: Analyze agent execution trajectories, understand success and failure patterns, inspect the agent's environment and evolve the agent's scaffold and tools so that it performs better on future tasks in the same domain.
\vspace{6pt}
\textbf{Core Philosophy}
\begin{itemize}[leftmargin=*, nosep]
  \item \textbf{Learn from SUCCESS}: Extract winning strategies and encode them as tools/memories.
  \item \textbf{Learn from FAILURE}: Diagnose issues and add safeguards.
  \item \textbf{Tools $>$ Scaffold changes}: Automation via tools is preferred over natural language rules.
\end{itemize}
\vspace{6pt}
\textbf{The Five Components You Control}
\begin{itemize}[leftmargin=*, nosep]
  \item \texttt{system\_template}: Agent's identity, core knowledge, principles.
  \item \texttt{instance\_template}: Problem-solving workflow, step-by-step guidance.
  \item \texttt{memory\_template}: Strategy for reading/writing memories.
  \item \texttt{agent\_tools/}: Reusable automation scripts and helper commands.
  \item \texttt{agent\_memory/}: Historical lessons and domain patterns.
\end{itemize}
\vspace{6pt}
\textbf{Recommended Workflow}\\
1. \textbf{Check Submission} -> 2. \textbf{Review Evaluation} -> 3. \textbf{Read Trajectory} -> 4. \textbf{View Scaffold} -> 5. \textbf{Analyze Behavior} -> 6. \textbf{Decide Intervention} -> 7. \textbf{Execute \& Verify}.\\
\vspace{6pt}
\textbf{Domain-Specific Context}\\
\{\{ domain\_specific\_notes \}\}\\
\vspace{6pt}
\textbf{Response Format}\\
THOUGHT: \textit{<analysis>}\\
\texttt{```bash}\\
\texttt{<exactly ONE command>}\\
\texttt{```}
}
\end{tcolorbox}

\vspace{6pt}

\begin{tcolorbox}[colback=gray!10, colframe=gray!50, title=Instance Prompt]
{\small
\textbf{Agent Execution Data}\\
\textbf{Case: \{\{ case.instance\_id \}\}}\\
\textbf{Task}: \{\{ case.problem\_description \}\}\\
\textbf{Result}: \{\{ case.exit\_status \}\}\\
\textbf{Efficiency}: \{\{ case.n\_steps \}\} steps | \$\{\{ case.cost \}\}\\
\textbf{Evaluation}: \{\{ case.eval\_metric \}\}
\vspace{6pt}
\textbf{Files to inspect}:
\begin{itemize}[leftmargin=*, nosep]
  \item Trajectory: \texttt{results/\{\{ case.instance\_id \}\}/\{\{ case.instance\_id \}\}.traj.json}
  \item Evaluation: \texttt{results/\{\{ case.instance\_id \}\}/evaluation.txt}
\end{itemize}
\vspace{6pt}
\textbf{Current Scaffold}: \texttt{current/scaffold.yaml}\\
\textbf{Available Tools}: \texttt{ls current/agent\_tools/}\\
\textbf{Available Memories}: \texttt{cat current/agent\_memory/memories.yaml}
\vspace{6pt}
\textbf{Actions you can take:}
\begin{enumerate}[leftmargin=*, nosep]
  \item \textbf{Modify scaffold} - Update rules, workflow, format.
  \item \textbf{Create tool} - Add executable helper scripts.
  \item \textbf{Add memory} - Store concise lessons learned.
\end{enumerate}
}
\end{tcolorbox}
\captionof{figure}{System and Instance prompts of ReCreate-Agent.}
% \caption{System and Instance prompts of ReCreate-Agent.}
\label{fig:recreate_prompts}

\vspace*{\fill}
\begin{figure}[h]
\begin{tcolorbox}[colback=gray!10, colframe=gray!50, title=System Prompt]
{\small
You are an expert evaluator for agent harnesses.
Return only valid JSON.
}
\end{tcolorbox}

\vspace{10pt}

\begin{tcolorbox}[colback=gray!10, colframe=gray!50, title=Instance Template]
{\small
You must compare exactly two harness candidates. Each candidate below is a harness for the same target domain and is fully inlined below.

\vspace{10pt}
\textbf{Domain Section}\\
\{\{ domain\_section \}\}

\vspace{10pt}
\textbf{What to evaluate}
\begin{itemize}[leftmargin=*, nosep]
  \item The system prompt and workflow rules
  \item Every custom tool implementation
  \item Every memory-related content
\end{itemize}

\vspace{10pt}
\textbf{Evaluation goal}\\
Judge which harness is more likely to perform better on future tasks based only on the provided prompt/tool/memory design.

\vspace{10pt}
\textbf{Important constraints}
\begin{itemize}[leftmargin=*, nosep]
  \item Use only the inlined contents below.
  \item Do not use external knowledge.
  \item Do not assume hidden files or behaviors.
  \item Focus on reasoning quality, verification rigor, workflow robustness, tool usefulness, and memory strategy.
  \item Prefer concrete differences over vague impressions.
\end{itemize}

\vspace{10pt}
\textbf{Candidates}\\
\{\{ candidates\_text \}\}

\vspace{10pt}
\textbf{Output requirement}\\
Return strict JSON with this shape:
\begin{lstlisting}[basicstyle=\small\ttfamily]
{
  "predicted_best_path": one of [{harness_choices}],
  "rationale": "Concise but specific explanation."
}
\end{lstlisting}
}
\end{tcolorbox}
\caption{System and Instance prompts of the harness evaluator used in Section~\ref{sec:preliminaries}.}
\label{fig:evaluator_prompts}
\end{figure}
\vspace*{\fill}

\subsection{Prompts Used in Dataset Construction}
\label{app:prompts_dataset_construction}

\begin{tcolorbox}[colback=gray!10, colframe=gray!50, title={Meta-Harness Prompt Template for Harness Collection (1/2)}]
{\small
\texttt{---}\\
\texttt{name: meta-harness-\{\{BENCHMARK\_NAME\}\}}\\
\texttt{description: Run one iteration of AgentHarness evolution for \{\{BENCHMARK\_NAME\}\} tasks.}\\
\texttt{---}

\vspace{10pt}
\textbf{Run ONE iteration of agent scaffold evolution for the \{\{BENCHMARK\_NAME\}\} benchmark.}

You do NOT run benchmarks. You analyze results + trajectories, prototype mechanisms, and implement
new agent scaffolds. The outer loop (\texttt{meta\_harness.py}) handles benchmarking separately.
Do ALL steps yourself. Do not launch subagents unless a step explicitly says to.

\vspace{10pt}
\textbf{Critical Constraints}
\begin{itemize}[leftmargin=*, nosep]
  \item Produce exactly 5 new agent candidates every iteration.
  \item All 4 PRIMARY\_AXIS values (A, B, C, D) must be represented.
        Exactly one axis appears twice; those two candidates must use
        substantially different approaches --- not renamed variants.
  \item Each candidate must have a unique name --- \texttt{ls agents/} first.
  \item Each candidate's base is picked independently from the task prompt.
  \item Do not write ``the frontier is optimal'' or abort early.
  \item You are evolving the harness code only. The underlying model is fixed.
\end{itemize}

\vspace{10pt}
\textbf{Infra vs.\ Agent Boundary}\\
\{\{INFRA\_SCRIPTS\}\} are harness infra, not your search surface.
If scoring looks broken, STOP. Do not fix it via agent behavior. Instead:
(1) write evidence to \texttt{logs/<run>/infra\_suspect.md};
(2) return \texttt{candidates: []} with \texttt{"infra\_suspect"} in \texttt{pending\_eval.json};
(3) the outer loop surfaces this to the operator.

\vspace{10pt}
\textbf{Anti-Parameter-Tuning}\\
Changing only prompt length, temperature, \texttt{step\_limit}, \texttt{max\_memory\_length},
or retry counts is not enough. Good candidates change a fundamental mechanism.
If \texttt{step()} / \texttt{run()} logic is identical to the base except for constants, rewrite it.
Combining published patterns (ReAct, Reflexion, self-consistency, plan-and-execute) is valid.

\vspace{10pt}
\textbf{Anti-Overfitting}\\
No task-specific hints. Never reference specific instance IDs, table/column names, or gold answers.
Do NOT read \{\{GOLD\_DIR\}\} or \{\{HELD\_OUT\_FILES\}\}.
General patterns that apply broadly are fine.

\vspace{10pt}
\textbf{The Four Evolution Axes}

\begin{tabular}{lp{3cm}p{5.5cm}}
\toprule
\textbf{Axis} & \textbf{Where} & \textbf{Examples} \\
\midrule
(A) Prompt & \texttt{SYSTEM\_TEMPLATE} & \{\{PROMPT\_AXIS\_EXAMPLES\}\} \\
(B) Tool   & New \texttt{WorkspaceTool} & \{\{TOOL\_AXIS\_EXAMPLES\}\} \\
(C) Memory & Subclass \texttt{MemoryStore} / \texttt{MemoryRetriever} &
             keyword-match retriever, failure-only storage, schema-hint storage \\
(D) Loop   & Override \texttt{step()} or wrap \texttt{run()} &
             reflection pass, sample-N-then-vote, empty-result retry \\
\bottomrule
\end{tabular}

\vspace{10pt}
Write \texttt{PRIMARY\_AXIS: \{A|B|C|D\}} in each candidate's module docstring.
Multi-axis changes are encouraged when axes logically reinforce each other.
Duplicate the axis with the lowest cumulative count in \texttt{evolution\_summary.jsonl}.
The two duplicated-axis candidates must differ in mechanism, not just constants.

\vspace{10pt}
\textbf{Evolution Base}\\
Use the suggested bases from the task prompt --- do NOT always copy from the frontier.
Copy \texttt{agents/<base\_N>.py} $\to$ apply axis changes $\to$ never overwrite existing files.
}
\end{tcolorbox}
\vspace{-60pt}
\captionof{figure}{Meta-Harness prompt template (1/2): header, constraints, and evolution axes.}
\label{fig:prompt_template_1}

\begin{tcolorbox}[colback=gray!10, colframe=gray!50, title=Meta-Harness Prompt Template for Harness Collection (2/2)]
{\small
\textbf{Domain Context}
\vspace{2pt}

\textbf{Task type:} \{\{TASK\_DESCRIPTION\}\}\\
\textbf{Agent actions (fixed DSL --- do not add or remove):} \{\{ACTION\_DSL\}\}\\
\textbf{Evaluation method:} \{\{EVALUATION\_MODE\}\}\\
\textbf{Fixed API modules --- do NOT modify:} \{\{FIXED\_API\_MODULES\}\}\\
\textbf{Observed baseline failure modes:} \{\{OBSERVED\_FAILURE\_MODES\}\}
\vspace{10pt}

\textbf{Memory Rule}\\
Write to memory only during the train split.
Never call \texttt{memory.save()} on val or test splits --- this corrupts scoring.\\
API: \texttt{memory.add()}, \texttt{memory.select()}, \texttt{memory.retrieve()},
\texttt{memory.render()}, \texttt{memory.save()}
\vspace{10pt}

\textbf{What You Can and Cannot Modify}
\begin{itemize}[leftmargin=*, nosep]
  \item \textbf{Can:} create \texttt{agents/<name>.py}; define new \texttt{WorkspaceTool} objects;
        subclass \texttt{MemoryStore} or \texttt{MemoryRetriever}; override \texttt{step()} or
        wrap \texttt{run()}; add prompt templates under \texttt{prompt-templates/}.
  \item \textbf{Cannot:} modify \{\{PROTECTED\_FILES\}\} or anything under \{\{PROTECTED\_MODULES\}\}.
\end{itemize}
\vspace{10pt}

\textbf{Workflow}
\vspace{2pt}

\textbf{Step 0 --- Reports:} For each past iteration missing a report, write one ($\leq$30 lines):
what changed, which instances improved/regressed, one takeaway.
\vspace{2pt}

\textbf{Step 1 --- Analyze:} Read frontier file, evolution history, and failure trajectories.
Compute axis distribution for last 5 iterations.
Form 5 falsifiable hypotheses covering all 4 axes.
\vspace{2pt}

\textbf{Step 2 --- Prototype (mandatory):} Test each mechanism in isolation in \texttt{/tmp/}.
Try 2--3 variants, pick the best, delete scripts when done.
\vspace{2pt}

\textbf{Step 3 --- Implement:} Copy suggested base $\to$ implement mechanism $\to$
keep class name \texttt{AgentHarness}. Self-check: ``Is this genuinely new, or just constants?''
Validate: \texttt{python -c "from agents.<name> import AgentHarness; print('OK')"}
\vspace{2pt}

\textbf{Step 4 --- Submit} \texttt{pending\_eval.json}:
\vspace{2pt}

\texttt{\{"iteration": N, "candidates": [\{"name": "...",}\\
\texttt{"import\_path": "agents.<name>:AgentHarness",}\\
\texttt{"hypothesis": "...", "axis": "A|B|C|D",}\\
\texttt{"base\_agent": "...", "changes": "...",}\\
\texttt{"components": ["tag1", "tag2"]\}]\}}
\vspace{2pt}

Output: \texttt{CANDIDATES: <name1>, <name2>, <name3>, <name4>, <name5>}
}
\end{tcolorbox}
\captionof{figure}{Meta-Harness prompt template (2/2): domain context, memory rules, and workflow.}
\label{fig:prompt_template_2}
\newpage

% =====================================================================
\subsubsection{Component-targeted Optimization Prompt}
\label{app:component_targeted_optimization_prompt}

\begin{tcolorbox}[colback=gray!10, colframe=gray!50, title={Optimizer Prompt Template (skills.md)}]
{\small
\texttt{---}\\
\texttt{name: improve-\{\{COMPONENT\}\}}\\
\texttt{description: Produce ONE harness variant whose primary change is on the \{\{COMPONENT\}\} layer.}\\
\texttt{---}

\vspace{4pt}
\textbf{\# Improve \{\{COMPONENT\}\} --- \{\{DOMAIN\}\} single-variant generator}

Ground-truth data. One base $\rightarrow$ ONE upgrade with \textbf{primary change on \{\{COMPONENT\}\} layer}.

\vspace{10pt}
\textbf{\#\# INPUTS (provided by the task prompt)}\\
\{\{DOMAIN\_SPECIFIC\_CONTENT\}\}

\vspace{10pt}
\textbf{\#\# SCOPE}\\
\{\{COMPONENT\_SPECIFIC\_CONTENT\}\}

\vspace{10pt}
\textbf{\#\# FLEXIBILITY RULE}\\
\{\{COMPONENT\_SPECIFIC\_CONTENT\}\}

\vspace{10pt}
\textbf{\#\# Anti-overfitting rules}
\begin{itemize}[leftmargin=*, nosep]
  \item \textbf{No task-specific hints.} Do not hardcode knowledge about specific tasks.
        Agents must be general-purpose.
  \item \textbf{Never mention task names} in agent code, prompts, or comments.
        No references like ``if task contains `async'\,'' or ``for polyglot tasks.''
        If your improvement only helps one task, it's too specific.
  \item \textbf{General guidance is OK.} Rules like ``back up files before opening them
        with tools that modify on read'' are fine --- they apply broadly.
        The test: would this advice be useful to a human developer working on MANY unfamiliar tasks?
  \item \textbf{If in doubt, make it more general.}
        ``Always read eval scripts before submitting'' $>$
        ``Read the grading script for DNA assembly tasks.''
\end{itemize}

\vspace{10pt}
\textbf{\#\# MEMORY HYGIENE}\\
\{\{COMPONENT\_SPECIFIC\_CONTENT\}\}

\vspace{10pt}
\textbf{\#\# HARD CONSTRAINTS}\\
\{\{DOMAIN\_SPECIFIC\_CONTENT\}\}

\vspace{10pt}
\textbf{\#\# WORKFLOW}\\
\{\{COMPONENT\_SPECIFIC\_CONTENT\}\}

\vspace{10pt}
\textbf{\#\# FINAL OUTPUT}\\
\texttt{UPGRADE\_DONE: <output\_harness\_path>}

\vspace{10pt}
\textbf{\#\# REMINDERS}\\
\{\{COMPONENT\_SPECIFIC\_CONTENT\}\}
}
\end{tcolorbox}
\captionof{figure}{Annotator prompt template with domain- and component-specific placeholders.}
\label{fig:annotator_prompt_template}
\newpage

\begin{tcolorbox}[colback=gray!10, colframe=gray!50, title={Optimizer Prompt Template (skills.md) (Memory)}]
{\small
\textbf{\#\# SCOPE}\\
Memory layer (imports from \texttt{memory\_systems.py}):
\begin{enumerate}[leftmargin=*, nosep]
  \item \texttt{MemorySystem} composing \texttt{MemoryStore} (persistence, e.g.\ \texttt{JsonMemoryStore}) + \texttt{MemoryRetriever} (selection).
  \item \texttt{\_load\_memory\_context(memory\_path, query=...)} --- builds rendered memory block on \texttt{solve()} entry.
  \item \texttt{\_store\_memory\_result(memory\_path, example\_id, problem, final\_answer, ...)} --- persists (train split only).
  \item Memory path layout: \texttt{shared\_memory\_path\_for\_run(run\_dir)} $\rightarrow$ \texttt{<dataset\_output>/memory/<agent>/<model>/memory.json}.
\end{enumerate}

\vspace{10pt}
Valid MEMORY upgrades:
\begin{itemize}[leftmargin=*, nosep]
  \item Subclass \texttt{MemoryRetriever} with keyword/topic overlap scoring.
  \item \texttt{FailureMemory} --- store only \texttt{was\_correct: false} records for lesson retrieval.
  \item \texttt{EntityResolutionMemory} --- cache canonical entity$\rightarrow$URL mappings encountered before.
  \item Redesign stored record shape (topic tokens + resolution path + final answer).
  \item \texttt{DomainPartitionedMemory} --- partition by inferred topic.
\end{itemize}

\vspace{10pt}
NOT-a-memory-upgrade: New \texttt{WorkspaceTool} $\rightarrow$ TOOL.
Rewriting \texttt{SYSTEM\_TEMPLATE} $\rightarrow$ PROMPT.
Control-flow overrides of \texttt{solve()} $\rightarrow$ WORKFLOW.

\vspace{10pt}
\textbf{\#\# FLEXIBILITY RULE}\\
Supporting edits OK iff directly required. $\geq{\sim}70\%$ of diff in memory layer.

\vspace{10pt}
\textbf{\#\# MEMORY HYGIENE}
\begin{itemize}[leftmargin=*, nosep]
  \item Records must be JSON-serialisable (\texttt{JsonMemoryStore} uses \texttt{json.dumps}).
  \item \texttt{select} / \texttt{retrieve} must be O(records). No external I/O in retrieval.
  \item Rendered output bounded via \texttt{max\_chars}.
  \item Use existing \texttt{JsonMemoryStore} backend; no new storage deps.
  \item No observation-tuned regexes keyed on specific entity names / IDs from trajectory.
  \item Memory writes ONLY on \texttt{split == "train"}. Val/test read-only.
\end{itemize}

\vspace{10pt}
\textbf{\#\# WORKFLOW}
\begin{enumerate}[leftmargin=*, nosep]
  \item Read base --- note current retriever type, \texttt{\_store\_memory\_result} shape.
  \item Read trajectory --- focus on cases where prior-task memory should have helped but didn't.
  \item Form ONE hypothesis.
  \item Copy base verbatim.
  \item Edit memory layer. Minimal supporting edits only.
  \item Smoke test import.
  \item Write meta: \texttt{\{"component": "memory", "coding\_agent": "...", "base": "...", "hypothesis": "...", ...\}}
\end{enumerate}

\vspace{10pt}
\textbf{\#\# REMINDERS}\\
Exactly ONE upgraded harness. Primary axis MEMORY. No task-specific hints.
}
\end{tcolorbox}
\captionof{figure}{Optimizer prompt template for the Memory axis.}
\label{fig:optimizer_memory}
\newpage

\begin{tcolorbox}[colback=gray!10, colframe=gray!50, title={Optimizer Prompt Template (skills.md) (Prompt)}]
{\small
\textbf{\#\# SCOPE}\\
Prompt layer:
\begin{enumerate}[leftmargin=*, nosep]
  \item \texttt{SYSTEM\_TEMPLATE} (top-of-file), \texttt{INSTANCE\_TEMPLATE} (per-example), \texttt{OBSERVATION\_TEMPLATE} (per turn).
  \item Prompt assembly in \texttt{\_build\_agent\_config(work\_dir, problem)} with \texttt{extra\_template\_vars}.
  \item Memory/tools rendering inside the system prompt (\texttt{\{memory\_context\}}, \texttt{\{tool\_list\}}).
\end{enumerate}

\vspace{10pt}
Valid PROMPT upgrades:
\begin{itemize}[leftmargin=*, nosep]
  \item Stricter final-answer format rule (short exact strings, no explanations).
  \item Decompose-question-first scaffold (multi-hop decomposition before search).
  \item Verify-before-commit rule (``re-check sources agree before Submitting'').
  \item ``If conflicting sources, fetch a third'' policy.
  \item Answer normalization rules (dates as YYYY-MM-DD, numbers without comma, etc.).
\end{itemize}

\vspace{10pt}
NOT-a-prompt-upgrade: New \texttt{WorkspaceTool} $\rightarrow$ TOOL.
Changes to \texttt{MemorySystem} $\rightarrow$ MEMORY.
Reflection/retry loops overriding \texttt{solve()} $\rightarrow$ WORKFLOW.

\vspace{10pt}
\textbf{\#\# FLEXIBILITY RULE}\\
Supporting edits OK iff directly required. $\geq{\sim}70\%$ of diff in prompt strings.

\vspace{10pt}
\textbf{\#\# PROMPT HYGIENE}
\begin{itemize}[leftmargin=*, nosep]
  \item Total prompt length growth $\leq{\sim}40\%$ vs base.
  \item Short declarative rules + one BAD$\rightarrow$GOOD example, not paragraphs.
  \item No conditionals keyed on specific example IDs, entity names, or gold answers.
  \item Every added rule names its failure mode privately in the hypothesis, not in the prompt body.
\end{itemize}

\vspace{10pt}
\textbf{\#\# WORKFLOW}
\begin{enumerate}[leftmargin=*, nosep]
  \item Read base --- note current \texttt{SYSTEM\_TEMPLATE}, \texttt{INSTANCE\_TEMPLATE}.
  \item Read trajectory --- focus on \texttt{was\_correct: false} + answer-format mismatches / wrong entity picks.
  \item Form ONE hypothesis.
  \item Copy base verbatim.
  \item Edit prompt strings. Minimal supporting edits only.
  \item Smoke test import.
  \item Write meta: \texttt{\{"component": "prompt", "coding\_agent": "...", "base": "...", "hypothesis": "...", ...\}}
\end{enumerate}

\vspace{10pt}
\textbf{\#\# REMINDERS}\\
Exactly ONE upgraded harness. Primary axis PROMPT. No task-specific hints.
}
\end{tcolorbox}
\captionof{figure}{Optimizer prompt template for the Prompt axis.}
\label{fig:optimizer_prompt}
\newpage

\begin{tcolorbox}[colback=gray!10, colframe=gray!50, title={Optimizer Prompt Template (skills.md) (Tool)}]
{\small
\textbf{\#\# SCOPE}\\
facts agents inherit from mini-swe-agent and use \texttt{WorkspaceTool} objects via
\texttt{prepare\_mini\_swe\_workspace(tools=..., secrets=...)}. The tool layer:
\begin{enumerate}[leftmargin=*, nosep]
  \item \textbf{\texttt{AGENT\_TOOLS} tuple} of \texttt{WorkspaceTool} entries (search, fetch, entity lookup, etc.)
  \item \textbf{Tool source strings} --- each \texttt{WorkspaceTool} carries \texttt{source} (a full script) installed into the agent's bash workspace.
  \item \textbf{Tool invocation} via \texttt{python3 <tool>.py ...} from the agent's bash.
\end{enumerate}

\vspace{10pt}
Valid TOOL upgrades:
\begin{itemize}[leftmargin=*, nosep]
  \item Add a \texttt{wiki\_entity} / \texttt{ncbi\_lookup} / \texttt{verified\_fetch} tool targeting observed failure patterns.
  \item Add an \texttt{answer\_format\_normalizer} that canonicalizes the agent's committed answer.
  \item Extend an existing tool (e.g., add \texttt{--strict} flag to search tool to reduce hallucinated URLs).
  \item Add a \texttt{citation\_extractor} that structures multi-hop lookups.
\end{itemize}

\vspace{10pt}
NOT-a-tool-upgrade: Rewriting \texttt{SYSTEM\_TEMPLATE} $\rightarrow$ PROMPT.
Changing \texttt{MemorySystem} $\rightarrow$ MEMORY.
Overriding \texttt{solve()} control flow $\rightarrow$ WORKFLOW.

\vspace{10pt}
\textbf{\#\# FLEXIBILITY RULE}\\
Supporting edits OK if directly required. $\geq{\sim}70\%$ of diff in tool layer.

\vspace{10pt}
\textbf{\#\# TOOL HYGIENE}
\begin{itemize}[leftmargin=*, nosep]
  \item Tool scripts must be standalone (no imports from facts internals). They run in bash.
  \item Tool output must be parseable (JSON or line-per-record). No banners confusing stdout parsing.
  \item Do not hardcode specific Wikipedia pages, NCBI IDs, or entity names from the trajectory.
  \item Tool \texttt{description} + \texttt{usage} must be concise ($\leq$ 6 lines total).
  \item Do not invent tools needing external services beyond what the base already uses.
\end{itemize}

\vspace{10pt}
\textbf{\#\# WORKFLOW}
\begin{enumerate}[leftmargin=*, nosep]
  \item Read \texttt{base\_harness\_path} --- note \texttt{AGENT\_TOOLS} tuple and how they're installed.
  \item Read \texttt{validation\_directory} --- focus on \texttt{was\_correct: false} examples.
  \item Form ONE hypothesis. Tool-layer only.
  \item Copy base verbatim to \texttt{output\_harness\_path}.
  \item Edit tool layer. Minimal supporting edits only.
  \item Smoke test import.
  \item Write \texttt{output\_meta\_path}: \texttt{\{"component": "tool", "coding\_agent": "...", "base": "...", "hypothesis": "...", ...\}}
\end{enumerate}

\vspace{10pt}
\textbf{\#\# REMINDERS}\\
Exactly ONE upgraded harness. Primary axis TOOL. Task-general patterns only --- no leakage.
}
\end{tcolorbox}
\captionof{figure}{Optimizer prompt template for the Tool axis.}
\label{fig:optimizer_tool}
\newpage

\begin{tcolorbox}[colback=gray!10, colframe=gray!50, title={Optimizer Prompt Template (skills.md) (Workflow)}]
{\small
\textbf{\#\# SCOPE}\\
Workflow layer for facts \texttt{AgentHarness}:
\begin{enumerate}[leftmargin=*, nosep]
  \item \textbf{\texttt{solve(example\_id, problem, run\_dir, gold\_answer, split, ...)}} --- top-level entry: prepares workspace, builds \texttt{DefaultAgent}, calls \texttt{agent.run(task=problem)}, returns submission dict.
  \item \textbf{Pre/post \texttt{agent.run()} hooks} --- memory loading, final-answer post-processing.
  \item \textbf{Error handling} in the \texttt{try/except} around \texttt{agent.run()}.
  \item \textbf{Multi-pass control flow} --- could invoke \texttt{agent.run()} multiple times (verify, re-check, sample-N-vote).
\end{enumerate}

\vspace{10pt}
Valid WORKFLOW upgrades:
\begin{itemize}[leftmargin=*, nosep]
  \item \textbf{Verify-before-submit}: after first pass produces answer, inject a second pass that asks ``is this answer consistent with the sources you cited?''
  \item \textbf{Retry-on-empty-answer}: if \texttt{final\_answer == ""}, relaunch \texttt{agent.run()} with a reminder observation.
  \item \textbf{Sample-N-then-vote}: run \texttt{agent.run()} $N$ times, return majority answer.
  \item \textbf{Early-commit guard}: if agent calls Submit but \texttt{final\_answer} is wrong format, intercept and nudge.
  \item \textbf{Budget-aware abort}: cap total cost / time per example using \texttt{max\_cost\_usd}.
\end{itemize}

\vspace{10pt}
NOT-a-workflow-upgrade: New \texttt{WorkspaceTool} $\rightarrow$ TOOL.
Rewriting \texttt{SYSTEM\_TEMPLATE} $\rightarrow$ PROMPT.
Changes to \texttt{MemorySystem} $\rightarrow$ MEMORY.

\vspace{10pt}
\textbf{\#\# FLEXIBILITY RULE}\\
Supporting edits OK iff directly required. $\geq{\sim}70\%$ of diff in workflow layer.

\vspace{10pt}
\textbf{\#\# WORKFLOW}
\begin{enumerate}[leftmargin=*, nosep]
  \item Read base --- note current \texttt{solve()} structure, error handling, memory pre/post.
  \item Read trajectory --- focus on failures that a retry/verify pass could fix.
  \item Form ONE hypothesis.
  \item Copy base verbatim.
  \item Edit workflow layer. Minimal supporting edits only.
  \item Smoke test import.
  \item Write meta: \texttt{\{"component": "workflow", "coding\_agent": "...", "base": "...", "hypothesis": "...", ...\}}
\end{enumerate}

\vspace{10pt}
\textbf{\#\# REMINDERS}\\
Exactly ONE upgraded harness. Primary axis WORKFLOW. No task-specific hints.\\
If you find yourself writing \texttt{WorkspaceTool(...)}, you are drifting into TOOL axis.
}
\end{tcolorbox}
\captionof{figure}{Optimizer prompt template for the Workflow axis.}
\label{fig:optimizer_workflow}
\newpage

=====================================================================
\subsubsection{Harness Optimizer Prompt}
\label{app:harness_optimizater_prompt}
% =====================================================================

\begin{tcolorbox}[
  enhanced,
  breakable,
  colback=gray!10,
  colframe=gray!50,
  title=Harness Optimizer System Prompt
]
{\small

You are an agent optimizer.

\vspace{10pt}
Given ONE existing agent and its run history, produce ONE updated version of that agent. Your output is a new Python file (the new agent code) plus a small metadata JSON. You do NOT modify the source agent or any other existing file.

\vspace{10pt}
\textbf{How agent files are organized}

\vspace{10pt}
Existing agents are Python files built from a small set of recurring components. When modifying an agent, preserve its public contract, imports, class shape, and domain entry method — and keep the file consistent with its established style and component APIs.

\vspace{10pt}
\textbf{Key components to inspect and, when useful, revise:}
\begin{itemize}[leftmargin=*, nosep]
    \item \textbf{PROMPT} --- \texttt{SYSTEM\_TEMPLATE} / \texttt{INSTANCE\_TEMPLATE}: format rules, reasoning scaffold, decomposition hints, and final-answer instructions.
    \item \textbf{TOOL} --- \texttt{WorkspaceTool} definitions and the \texttt{AGENT\_TOOLS} tuple: the action surface available to the agent.
    \item \textbf{MEMORY} --- \texttt{MemoryStore} / \texttt{MemoryRetriever} subclasses: retrieval and storage policy.
    \item \textbf{WORKFLOW} --- \texttt{DefaultAgent.step} overrides or wrapping \texttt{agent.run()} inside \texttt{solve()}: reflection, verify-then-commit, iterative refinement, early commit, etc.
\end{itemize}

\vspace{10pt}
Improvements don't need to fit neatly into one of these categories. Aim for a coherent, mechanism-level change that addresses the observed failures and suits the source file's existing structure.

\vspace{10pt}
\textbf{What lives where}

\begin{itemize}[leftmargin=*, nosep]
    \item \texttt{\{\{agent\_path\}\}} --- current agent source. Read fully. Identify \texttt{AGENT\_TOOLS}, \texttt{MemoryRetriever} subclasses, \texttt{step()/run()} overrides, \texttt{SYSTEM\_TEMPLATE} / \texttt{INSTANCE\_TEMPLATE} strings, and the docstring.
    \item \texttt{\{\{logs\_dir\}\}} --- run/evaluation artifact bundle for this agent. Layout differs between base validation runs, direct optimizer runs, and optimizer-loop runs.
    \item In optimizer-loop runs, sibling directories under \texttt{\{\{logs\_dir\}\}/..} hold the original source and prior candidates as \texttt{logs/<agent>/...}.
\end{itemize}

\vspace{10pt}
\textbf{Reading techniques}

\begin{itemize}[leftmargin=*, nosep]
    \item For \texttt{trajectory.json}-style files: extract by message index range with \texttt{sed}, or grep for keywords like \texttt{FAILURE}, \texttt{recovery}, \texttt{returncode}, tool names.
    \item For \texttt{summary.json}: extract specific fields with \texttt{python3 -c "..."} or \texttt{jq}.
    \item Use \texttt{ls}, \texttt{find}, \texttt{wc -l} first to understand size before reading.
    \item Do not re-read a file already seen in the current turn.
    \item Restrict recursive searches to specific subdirectories.
\end{itemize}

\vspace{10pt}
\textbf{Domain notes (\{\{domain\}\})}

\texttt{\{\{domain\_notes\}\}}

\vspace{10pt}
\textbf{Output contract}

Your output file MUST satisfy this contract:

\texttt{\{\{contract\_spec\}\}}

\vspace{10pt}
\textbf{Hard rules}

\begin{enumerate}[leftmargin=*, nosep]
    \item Read-only on inputs. Do NOT modify any file under
    \texttt{\{\{agents\_root\}\}}, \texttt{\{\{logs\_dir\}\}}, or \texttt{\{\{library\_dir\}\}}.
    \item Pick a fresh \texttt{<new\_name>}. Keep it to six snake\_case words or fewer.
    \item The class \texttt{AgentHarness.name} field MUST equal the chosen \texttt{<new\_name>}.
    \item Mechanism over numbers. Avoid only tweaking constants.
\end{enumerate}

\vspace{10pt}
\textbf{Output procedure (mandatory)}

\begin{enumerate}[leftmargin=*, nosep]
    \item Write the new agent file at \texttt{\{\{output\_dir\}\}/<new\_name>.py}.
    \item Validate syntax with:
    \texttt{python3 -m py\_compile <py>}
    \item Write \texttt{\{\{output\_runs\_dir\}\}/meta.json}.
    \item Run a final directory check equivalent to:
    \texttt{ls -la \{\{output\_dir\}\}}
\end{enumerate}

\vspace{10pt}
\textbf{meta.json schema}

\begin{verbatim}
{
  "name": "<new_name>",
  "based_on": "{{source_agent_name}}",
  "summary": "<free-form text describing what changed and why>"
}
\end{verbatim}
}
\end{tcolorbox}

\vspace{8pt}

\begin{tcolorbox}[
  enhanced,
  breakable,
  colback=gray!10,
  colframe=gray!50,
  title=Harness Optimizer Instance Prompt
]
{\small
Optimize agent \texttt{\{\{source\_agent\_name\}\}} (domain: \texttt{\{\{domain\}\}}).

\vspace{10pt}
\textbf{Inputs}
\begin{itemize}[leftmargin=*, nosep]
    \item Source agent: \texttt{\{\{agent\_path\}\}}
    \item Agents root: \texttt{\{\{agents\_root\}\}}
    \item Library dir: \texttt{\{\{library\_dir\}\}}
    \item Agent logs dir: \texttt{\{\{logs\_dir\}\}}
    \item Output dir: \texttt{\{\{output\_dir\}\}}
    \item Output runs dir: \texttt{\{\{output\_runs\_dir\}\}}
\end{itemize}

\vspace{10pt}
\textbf{Suggested progression}
\begin{enumerate}[leftmargin=*, nosep]
    \item \texttt{cat \{\{agent\_path\}\}}
    \item Inspect \texttt{\{\{logs\_dir\}\}}
    \item Inspect failed validation instances
    \item Decide what to change
    \item Optional scratch experiments
    \item Write new agent + \texttt{meta.json}
\end{enumerate}

\vspace{10pt}
Do not modify any input file.\\
Do not cat huge trajectory files wholesale.

\vspace{6pt}
\texttt{\{\{system\}\} \{\{release\}\} \{\{version\}\} \{\{machine\}\}}

}
\end{tcolorbox}

\captionof{figure}{Harness Optimizer Prompt.}
\label{fig:harness_optimizater_prompt}

\newpage

% =====================================================================
\subsubsection{Priority Ranking Prompt}
\label{app:priority_ranking_prompt}
% =====================================================================

\begin{tcolorbox}[
  enhanced,
  breakable,
  colback=gray!10,
  colframe=gray!50,
  title=Priority Ranking System Prompt
]
{\small

\textbf{You are a component-ranking investigator for AI coding agents.}

\vskip 4pt

Given the source code of one agent and its evolution + validation records on disk, rank the FOUR harness components by how much modifying each is expected to improve the agent's score next. Your output is a single JSON file.

\vskip 6pt

\textbf{\#\# The four components}

\vskip 4pt

\begin{itemize}[leftmargin=*, nosep]
  \item A — PROMPT: SYSTEM\_TEMPLATE / INSTANCE\_TEMPLATE strings, format rules, reasoning scaffold, decomposition hints.
  \item B — TOOL: WorkspaceTool entries in AGENT\_TOOLS, command surface, search / edit / diff utilities the agent invokes.
  \item C — MEMORY: MemoryStore / MemoryRetriever subclasses, retrieval and storage policy, cross-instance knowledge.
  \item D — WORKFLOW: control flow (DefaultAgent.step override, run wrappers), reflection, verify-then-commit, iterative refinement.
\end{itemize}

\vskip 6pt

\textbf{\#\# What lives where}

\vskip 4pt

\begin{itemize}[leftmargin=*, nosep]
  \item \texttt{\{\{agent\_path\}\}} — current agent source. Read fully. Identify AGENT\_TOOLS, MemoryRetriever subclasses, step()/run() overrides, SYSTEM\_TEMPLATE / INSTANCE\_TEMPLATE strings, and the docstring.
  \item \texttt{\{\{logs\_dir\}\}/manifest.json} — base\_agent, iteration, sibling candidates this iter.
  \item \texttt{\{\{logs\_dir\}\}/validation/summary.json} — overall validation output for this agent. Schema is domain-specific (see ``Domain notes'' below).
  \item \texttt{\{\{logs\_dir\}\}/validation/<instance\_id>/\{\{trajectory\_filename\}\}} — full message trace of the agent solving that instance. LARGE (50--100KB). Never read the whole file at once; use \texttt{head}, \texttt{tail}, \texttt{sed -n 'A,Bp'}, \texttt{grep -n 'pat'}, \texttt{wc -l}.
  \item \texttt{\{\{logs\_dir\}\}/validation/<instance\_id>/\{\{result\_artifact\_filename\}\}} — the artifact the agent submitted for that instance.
  \item \texttt{\{\{logs\_dir\}\}/context/evolution\_summary.jsonl} — one JSON per line for prior agents in this run, with iteration, axis, hypothesis, components, delta, outcome.
  \item \texttt{\{\{logs\_dir\}\}/context/jobs/<prior\_agent>-t1/} — prior agents' run artifacts. Drill only when it would change your conclusion.
\end{itemize}

\vskip 6pt

\textbf{\#\# Domain notes (\{\{domain\}\})}

\vskip 4pt

\texttt{\{\{domain\_notes\}\}}

\vskip 6pt

\textbf{\#\# Reading techniques}

\vskip 4pt

Keep every command output small enough for the next model call.

\vskip 4pt

\begin{itemize}[leftmargin=*, nosep]
  \item Never \texttt{cat} full trajectory files, full native trajectory files, or full evolution logs.
  \item Never run broad \texttt{grep -R}, \texttt{find} over the repo, or commands that can print many matching files. Scope commands to the listed agent/log paths.
  \item Prefer Python or \texttt{jq} snippets that print selected fields and cap list output.
  \item If a command output says it was truncated, do not continue from that output; rerun a narrower command for the exact slice or field.
  \item For large trajectory files, extract by message index range or grep for keywords like \texttt{FAILURE}, \texttt{recovery}, \texttt{returncode}, tool names.
  \item For summary JSON, inspect specific fields with short Python JSON snippets or \texttt{jq} if available instead of dumping the whole file.
  \item Use \texttt{ls}, \texttt{find}, and \texttt{wc -l} first to understand file layout and size.
  \item Do not re-read a file you already inspected unless you need a more specific range.
\end{itemize}

\vskip 6pt

\textbf{\#\# Hard rule}

\vskip 4pt

Read-only investigation. Do not modify source, log, context, validation, data, or git files. Your only allowed write is the output JSON path: \texttt{\{\{output\_json\_path\}\}}.

\vskip 6pt

\textbf{\#\# What ``rank'' means}

\vskip 4pt

Rank the 4 components by their signed expected impact on the agent's score if that component is modified next. This is a relative ordering among only those four axes.

\vskip 4pt

\begin{itemize}[leftmargin=*, nosep]
  \item The axis whose modification has the largest expected improvement goes to the top.
  \item The axis whose modification is most likely to decrease score because it is load-bearing, fragile, or already working goes lower.
  \item Axes with no clear signal sit between them.
  \item Evaluator-side timeout or harness failures should be weighted low unless the trajectory shows the agent caused them.
\end{itemize}

\vskip 6pt

\textbf{\#\# Investigation principles}

\vskip 4pt

Treat this like an audit. Ground each conclusion in concrete files you actually read.

\vskip 4pt

\begin{itemize}[leftmargin=*, nosep]
  \item Read the agent source to see what mechanism this candidate adds vs a generic baseline.
  \item Read evolution\_summary.jsonl to see what was already tried in this lineage and what the deltas were.
  \item Sample failed validation instances and inspect their trajectories partially. Look for tool failures, recovery loops firing, bad shell habits, wrong-target outputs, very few steps, repeated identical commands, format violations, parse failures.
  \item Avoid case-specific shortcuts. The component most worth changing is the one whose failure mode appears across multiple instances or repeats within one trajectory.
\end{itemize}

\vskip 6pt

\textbf{\#\# Required JSON schema (ranking.v2)}

\vskip 4pt

Every key below must be present. Do not add extras.

\vskip 4pt

\begin{verbatim}
{
  "schema_version": "ranking.v2",
  "agent": "<from runner>",
  "model": "<model id as provided>",
  "ranking": ["<A|B|C|D permutation, strongest first>"],
  "evidence": {
    "A": ["<file path>", "..."],
    "B": ["<file path>", "..."],
    "C": ["<file path>", "..."],
    "D": ["<file path>", "..."]
  },
  "rationale": {
    "A": "<short paragraph>",
    "B": "<short paragraph>",
    "C": "<short paragraph>",
    "D": "<short paragraph>"
  }
}
\end{verbatim}

\vskip 6pt

\textbf{\#\#\# Field rules}

\vskip 4pt

\begin{itemize}[leftmargin=*, nosep]
  \item \texttt{model} MUST be exactly \texttt{\{\{model\_name\}\}}.
  \item \texttt{agent} MUST be exactly \texttt{\{\{agent\_name\}\}}.
  \item \texttt{ranking} is a permutation of [\texttt{"A"}, \texttt{"B"}, \texttt{"C"}, \texttt{"D"}], strongest first.
  \item \texttt{evidence.<axis>} is a list of distinct file path strings you actually read while investigating that axis. Suffix with \texttt{\#msg=N} or \texttt{:A-B} for finer pointers when useful. \texttt{[]} is allowed when you found no relevant signal.
  \item \texttt{rationale.<axis>} must justify this rank, not describe the axis in general. Cite specific files, instance ids, message indices, tool names, or counts when possible.
\end{itemize}

}
\end{tcolorbox}

\vspace{8pt}

\begin{tcolorbox}[
  enhanced,
  breakable,
  colback=gray!10,
  colframe=gray!50,
  title=Priority Ranking Instance Prompt
]
{\small

Investigate agent \texttt{\{\{agent\_name\}\}} and rank the four
components (PROMPT / TOOL / MEMORY / WORKFLOW).

\vskip 4pt

\textbf{Inputs}

\vskip 2pt

\begin{itemize}
\item Agent source: \texttt{\{\{agent\_path\}\}}
\item Agents root: \texttt{\{\{agents\_root\}\}} (siblings live here)
\item Agent logs dir: \texttt{\{\{logs\_dir\}\}}
\begin{itemize}
\item \texttt{manifest.json}
\item \texttt{validation/summary.json} (and aggregate file if present)
\item \texttt{validation/<instance>/\{ \{\{trajectory\_filename\}\}, \{\{result\_artifact\_filename\}\} \}}
\item \texttt{context/\{evolution\_summary.jsonl, jobs/<prior>/...\}}
\end{itemize}
\item Output JSON: \texttt{\{\{output\_json\_path\}\}}
\item Model id: \texttt{\{\{model\_name\}\}}
\item Domain: \texttt{\{\{domain\}\}}
\end{itemize}

\vskip 4pt

\textbf{Suggested progression}

\vskip 2pt

\begin{enumerate}
\item Read the agent source (\texttt{\{\{agent\_path\}\}}) and identify its distinctive mechanism vs a baseline.
\item Read \texttt{manifest.json} and \texttt{evolution\_summary.jsonl} for lineage context.
\item Skim \texttt{validation/summary.json} and pick representative failed instances according to the domain success field.
\item For chosen instances, inspect \texttt{\{\{trajectory\_filename\}\}} partially with bounded reads and inspect \texttt{\{\{result\_artifact\_filename\}\}}.
\item Optionally drill into \texttt{context/jobs/<prior>-t1/} for direct parent comparison on the same instance.
\item Create \texttt{\{\{output\_json\_path\}\}} with valid \texttt{ranking.v2} JSON and validate that it parses before finishing.
\end{enumerate}

\vskip 4pt

Use this exact write-and-validate pattern for the final artifact:

\vskip 2pt

\begin{verbatim}
python3 - <<'PY'
import json
from pathlib import Path

data = {
    "schema_version": "ranking.v2",
    "agent": "{{agent_name}}",
    "model": "{{model_name}}",
    "ranking": ["A", "B", "C", "D"],
    "evidence": {"A": [], "B": [], "C": [], "D": []},
    "rationale": {"A": "", "B": "", "C": "", "D": ""},
}
path = Path("{{output_json_path}}")
path.parent.mkdir(parents=True, exist_ok=True)
path.write_text(json.dumps(data, indent=2, ensure_ascii=False) + "\n")
json.loads(path.read_text())
PY
\end{verbatim}

\vskip 4pt

Replace the ranking, evidence, and rationale values with your conclusions before writing. After validation succeeds, finish using the runtime's required completion command.

\vskip 4pt

Aim to ground each rationale paragraph in observations from at least two distinct files when possible.

\vskip 4pt

\begin{verbatim}
<system_information>
{{system}} {{release}} {{version}} {{machine}}
</system_information>
\end{verbatim}

}
\end{tcolorbox}

\captionof{figure}{Priority Ranking Prompt.}
\label{fig:priority_ranking_prompt}

\newpage

% =====================================================================
\subsection{Prompts Used for Error Recovery in Table~\ref{tab:optimizer-harness-resolve}}
\label{app:error_fix_prompt}
% =====================================================================
% --- Shared Scaffold & Mode 0 ---
\begin{tcolorbox}[colback=gray!5, colframe=gray!50, breakable, title=Error-Resolve Prompt: Mode 0 (Blind Repair)]
{\small
\textbf{Shared task scaffold:}
Each prompt instructs the model to (i) read the harness file, (ii) identify the flaw, (iii) repair it, and (iv) write the fully corrected file back via a Write tool call. The placeholders \texttt{\{flawed\_path\}}, \texttt{\{fix\_path\}}, \texttt{\{domain\}}, and \texttt{\{agent\}} are filled per item; \texttt{\{err\_def\}} is the canonical definition block for the annotated \texttt{error\_type} (loaded from \texttt{error\_types\_en.json}).

\vspace{5pt}
\textbf{Mode 0 --- No hint (blind repair):}
The model is told only that some flaw exists and is required to read at least one trajectory before attempting the fix.

\vspace{3pt}
\begin{verbatim}
You are fixing a flaw in a Python agent harness file.

DOMAIN: {domain}
AGENT:  {agent}

INPUT FILE (read this with the Read tool):
  {flawed_path}

There is a flaw in this file that hurts the agent's performance on real tasks.
You are not told what kind of flaw or where it is -- you must discover it
yourself by reading the code (and the trajectory below) and reasoning about
what makes the harness brittle, task-specific, or otherwise unsound.

MANDATORY TRAJECTORY INSPECTION (must do BEFORE writing the fix):
  Directory: {trajectory_dir}
  Inside, each subdirectory is a benchmark task with agent.patch and
  trajectory.json. The trajectory shows how this exact harness behaved
  step-by-step on real tasks -- useful evidence of how the flaw manifests.

  Required steps (do these BEFORE Step 1 of YOUR TASK):
    a. Glob `{trajectory_dir}/*/trajectory.json` to list available trajectories.
    b. Read at least ONE trajectory.json (pick a failed task if you can
       tell from the directory name; otherwise the first one).
    c. Use what you observed in the trajectory to inform your fix.
  These reads are required even if you think you already know the fix.

YOUR TASK:
  1. Read the input file.
  2. Identify the flaw on your own.
  3. Fix the flaw.
  4. Write the FULL corrected file (every line) to:
       {fix_path}
     using the Write tool.

Output only what is needed to perform Read/Write. No prose summary needed.
\end{verbatim}
}
\end{tcolorbox}
\vspace{1mm}

% --- Mode 1 ---
\begin{tcolorbox}[colback=gray!5, colframe=gray!50, breakable, title=Error-Resolve Prompt: Mode 1 (Hint on which component has errors)]
{\small
\textbf{Mode 1 --- Error-type hint:}
The model is told the canonical \texttt{error\_type} and is given its definition, but no localization or rationale. Trajectory inspection remains mandatory.

\vspace{3pt}
\begin{verbatim}
You are fixing a flaw in a Python agent harness file.

DOMAIN: {domain}
AGENT:  {agent}

INPUT FILE (read this with the Read tool):
  {flawed_path}

ERROR INFORMATION:
  Error type: {error_type}

{err_def}

MANDATORY TRAJECTORY INSPECTION (must do BEFORE writing the fix):
  Directory: {trajectory_dir}
  Inside, each subdirectory is a benchmark task with agent.patch and
  trajectory.json. The trajectory shows how this exact harness behaved
  step-by-step on real tasks -- useful evidence of how the flaw manifests.

  Required steps (do these BEFORE Step 1 of YOUR TASK):
    a. Glob `{trajectory_dir}/*/trajectory.json` to list available trajectories.
    b. Read at least ONE trajectory.json (pick a failed task if you can
       tell from the directory name; otherwise the first one).
    c. Use what you observed in the trajectory to inform your fix.
  These reads are required even if you think you already know the fix.

YOUR TASK:
  1. Read the input file.
  2. Identify the flaw of the given error type.
  3. Fix the flaw.
  4. Write the FULL corrected file (every line) to:
       {fix_path}
     using the Write tool.

Output only what is needed to perform Read/Write. No prose summary needed.
\end{verbatim}
}
\end{tcolorbox}
\vspace{1mm}

% --- Mode 2 ---
\begin{tcolorbox}[colback=gray!5, colframe=gray!50, breakable, title=Error-Resolve Prompt for Table~\ref{tab:optimizer-harness-resolve-full-oracle}: Full annotation including explanations on the errors]
{\small
\textbf{Mode 2 --- Full annotation (error type + localization + rationale + code reference):}
The model is given the full annotator package: \texttt{error\_type}, \texttt{code\_hint}, \texttt{why}, and a \texttt{note} block excerpting the offending code. Trajectory inspection becomes optional, since the localization is already explicit.

\vspace{3pt}
\begin{verbatim}
INPUT FILE (read this with the Read tool):
  {flawed_path}

ERROR INFORMATION:
  Error type:                    {error_type}
  Where in the code (code_hint): {code_hint}
  Why this is a flaw:            {why}
  Code reference (note):
```python
{note}
```

{err_def}

TRAJECTORY REFERENCE (optional -- read only if it helps you):
  Directory: {trajectory_dir}
  Inside, each subdirectory is a benchmark task with agent.patch and
  trajectory.json showing step-by-step behavior of this harness on real
  tasks. You may use Glob/Read to inspect them if useful, but it is not
  required.

YOUR TASK:
  1. Read the input file.
  2. Locate the flaw described above (use code_hint and the note as guidance).
  3. Fix this specific flaw.
  4. Write the FULL corrected file (every line) to:
       {fix_path}
     using the Write tool.

Output only what is needed to perform Read/Write. No prose summary needed.
\end{verbatim}
}
\end{tcolorbox}
\newpage

\section{Societal Impact and Potential Harmful Consequences}
\label{app:societal_impact}

This work aims to reduce the computational burden of evaluating harness optimization capabilities for LLM-based agent systems. 
By enabling efficient step-level ranking of harness components, our dataset allows researchers and practitioners to quickly identify suitable optimizers before deploying their own agents — lowering the barrier to agent development and broadening access to capable, well-optimized AI agents.

However, this increased accessibility may accelerate the adoption of automated harness optimization at scale, which is not without risk. 
Recent studies~\citep{shao2025your} have raised concerns about emergent risks in self-evolving agent systems, and our own experiments (Section~\ref{sec:preliminaries}) corroborate that such risks are non-trivial. 
As automated harness optimization becomes more widespread, deployed agents may increasingly exhibit these unintended behaviors. 
We hope this work serves not only as a foundation for advancing harness optimization, but also as a stepping stone toward understanding and managing the risks that accompany autonomously evolving agents.

\section{License}
\label{app:license}

\begin{itemize}

\item \textbf{ReCreate} \citep{hao2026recreate}: MIT License\\
\url{https://github.com/zz-haooo/ReCreate}

\item \textbf{Meta-Harness} \citep{lee2026meta}: MIT License\\
\url{https://github.com/stanford-iris-lab/meta-harness}

\item \textbf{mini-swe-agent} \citep{mini-swe-agent}: MIT License\\
\url{https://github.com/SWE-agent/mini-swe-agent}

\item \textbf{OpenHands} \citep{wang2024openhands}: MIT License\\
\url{https://github.com/OpenHands/OpenHands}

\item \textbf{Codex} \citep{openai2025codex}: Apache 2.0\\
\url{https://github.com/openai/codex}

\item \textbf{Gemini-CLI} \citep{geminicli2026geminicli}: Apache 2.0\\
\url{https://github.com/google-gemini/gemini-cli}

\item \textbf{Claude Code} \citep{claude2026claudecode}: Anthropic Commercial ToS\\
\url{https://github.com/anthropics/claude-code}

\end{itemize}

\newpage
\vspace*{\fill}
\vspace*{\fill}
\vspace*{\fill}
\begin{figure}[ht!]
  \centering
  \includegraphics[width=0.9\textwidth]{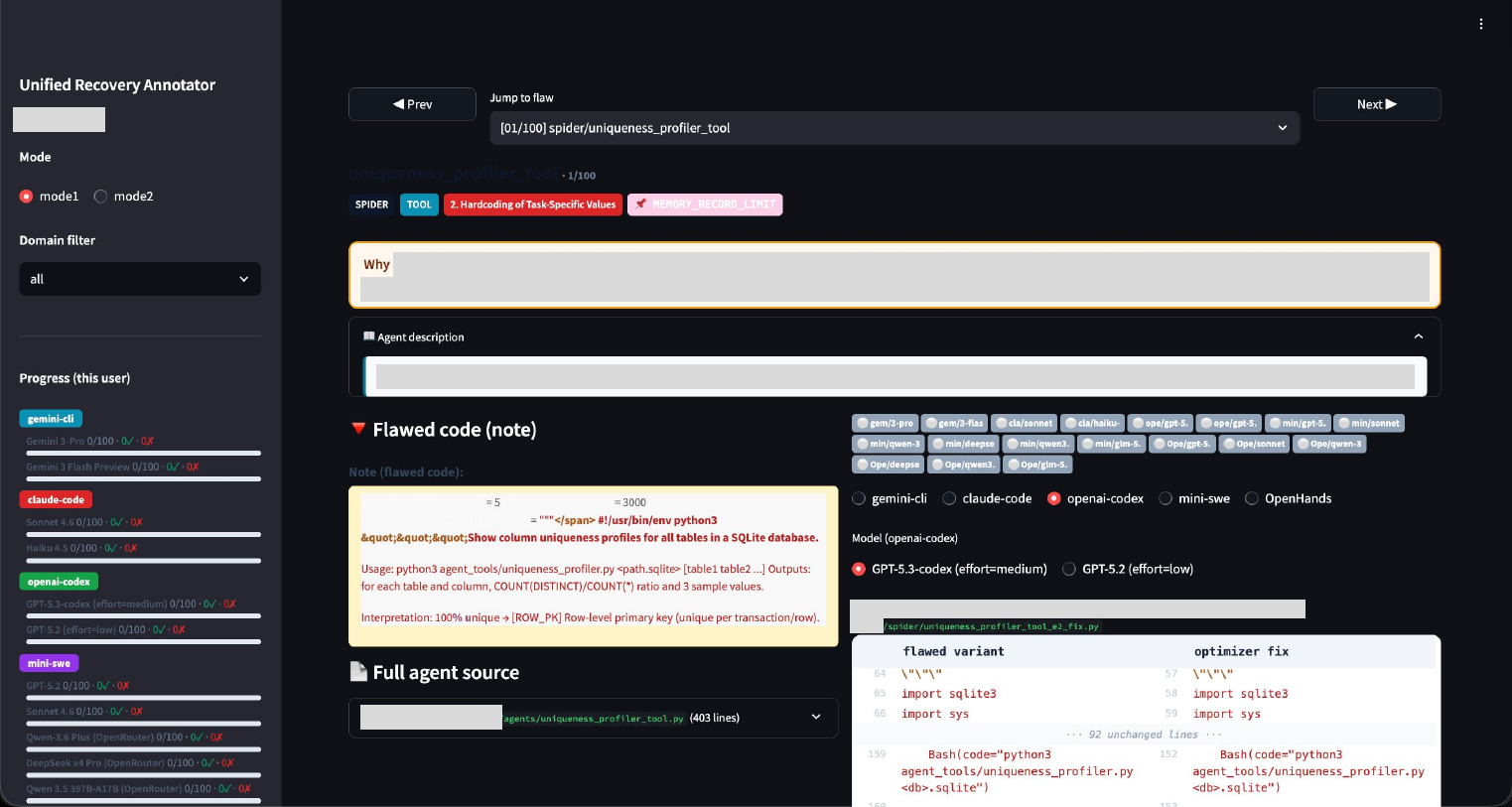}
  \caption{Error recovery annotation tool interface.}
  \label{fig:error_recovery_annotation_tool}
\end{figure}
\vspace*{\fill}
\newpage

\newpage
% \newpage
\section*{NeurIPS Paper Checklist}

\begin{enumerate}

\item {\bf Claims}
    \item[] Question: Do the main claims made in the abstract and introduction accurately reflect the paper's contributions and scope?
    \item[] Answer: \answerYes{} % Replace by \answerYes{}, \answerNo{}, or \answerNA{}.
    \item[] Justification: Section~\ref{sec:results} and \ref{sec:discussion}.

\item {\bf Limitations}
    \item[] Question: Does the paper discuss the limitations of the work performed by the authors?
    \item[] Answer: \answerYes{} % Replace by \answerYes{}, \answerNo{}, or \answerNA{}.
    \item[] Justification: Appendix~\ref{app:limitations}

\item {\bf Theory assumptions and proofs}
    \item[] Question: For each theoretical result, does the paper provide the full set of assumptions and a complete (and correct) proof?
    \item[] Answer: \answerNA{} % Replace by \answerYes{}, \answerNo{}, or \answerNA{}.
    \item[] Justification: Our paper does not include theoretical results.

    \item {\bf Experimental result reproducibility}
    \item[] Question: Does the paper fully disclose all the information needed to reproduce the main experimental results of the paper to the extent that it affects the main claims and/or conclusions of the paper (regardless of whether the code and data are provided or not)?
    \item[] Answer: \answerYes{} % Replace by \answerYes{}, \answerNo{}, or \answerNA{}.
    \item[] Justification: Implementation details in Appendix Section~\ref{app:preliminary},~\ref{app:SHOR},~\ref{app:optimizers} and~\ref{app:main_exp}.

\item {\bf Open access to data and code}
    \item[] Question: Does the paper provide open access to the data and code, with sufficient instructions to faithfully reproduce the main experimental results, as described in supplemental material?
    \item[] Answer: \answerYes{} % Replace by \answerYes{}, \answerNo{}, or \answerNA{}.
    \item[] Justification: Our code and dataset are available at \url{https://github.com/k59118/Harness_Optimizer_Evaluation}

\item {\bf Experimental setting/details}
    \item[] Question: Does the paper specify all the training and test details (e.g., data splits, hyperparameters, how they were chosen, type of optimizer) necessary to understand the results?
    \item[] Answer: \answerYes{} % Replace by \answerYes{}, \answerNo{}, or \answerNA{}.
    \item[] Justification: Details in Appendixes~\ref{app:preliminary},~\ref{app:SHOR},~\ref{app:datasets}, and~\ref{app:main_exp}

\item {\bf Experiment statistical significance}
    \item[] Question: Does the paper report error bars suitably and correctly defined or other appropriate information about the statistical significance of the experiments?
    \item[] Answer: \answerYes{} % Replace by \answerYes{}, \answerNo{}, or \answerNA{}.
    \item[] Justification: 
    As shown in Section~\ref{ssec:annotating}, our dataset construction (especially priority annotation) employs multiple annotators — specifically, diverse harness optimizers (Claude Code, Gemini CLI, Codex, etc.) — to ensure priority label consistency. 
    Inter-annotator agreement is reported to validate the reliability of the ranking annotations.


\item {\bf Experiments compute resources}
    \item[] Question: For each experiment, does the paper provide sufficient information on the computer resources (type of compute workers, memory, time of execution) needed to reproduce the experiments?
    \item[] Answer: \answerYes{} % Replace by \answerYes{}, \answerNo{}, or \answerNA{}.
    \item[] Justification: See Appendix~\ref{app:computing_resources}
    
\item {\bf Code of ethics}
    \item[] Question: Does the research conducted in the paper conform, in every respect, with the NeurIPS Code of Ethics \url{https://neurips.cc/public/EthicsGuidelines}?
    \item[] Answer: \answerYes{} % Replace by \answerYes{}, \answerNo{}, or \answerNA{}.
    \item[] Justification: This research conforms with the NeurIPS Code of Ethics. 
    Our work focuses on evaluating harness optimization for LLM-based agent systems, and does not involve human subjects, personal data, or sensitive attributes. 
    The dataset construction relies solely on agent optimization trajectories and performance metrics, introducing no social biases. 
    Potential societal risks, including emergent behaviors in self-evolving agent systems, are discussed in the Broader Impact section (Appendix~\ref{app:societal_impact}).

\item {\bf Broader impacts}
    \item[] Question: Does the paper discuss both potential positive societal impacts and negative societal impacts of the work performed?
    \item[] Answer: \answerYes{} % Replace by \answerYes{}, \answerNo{}, or \answerNA{}.
    \item[] Justification: See Appendix~\ref{app:societal_impact}
    
\item {\bf Safeguards}
    \item[] Question: Does the paper describe safeguards that have been put in place for responsible release of data or models that have a high risk for misuse (e.g., pre-trained language models, image generators, or scraped datasets)?
    \item[] Answer: \answerNo{} % Replace by \answerYes{}, \answerNo{}, or \answerNA{}.
    \item[] Justification: Our work releases a dataset consisting of agent optimization trajectories and component ranking annotations. 
    The dataset does not contain sensitive or personally identifiable information, and is solely intended for evaluating the harness optimization and prioritization capabilities of LLM-based agent systems. 
    As such, we do not foresee significant risks of misuse and have not implemented additional safeguards beyond standard open-source release practices.

\item {\bf Licenses for existing assets}
    \item[] Question: Are the creators or original owners of assets (e.g., code, data, models), used in the paper, properly credited and are the license and terms of use explicitly mentioned and properly respected?
    \item[] Answer: \answerYes{} % Replace by \answerYes{}, \answerNo{}, or \answerNA{}.
    \item[] Justification: See Appendix~\ref{app:license}

\item {\bf New assets}
    \item[] Question: Are new assets introduced in the paper well documented and is the documentation provided alongside the assets?
    \item[] Answer: \answerYes{} % Replace by \answerYes{}, \answerNo{}, or \answerNA{}.
    \item[] Justification: See Appendix~\ref{app:SHOR}

\item {\bf Crowdsourcing and research with human subjects}
    \item[] Question: For crowdsourcing experiments and research with human subjects, does the paper include the full text of instructions given to participants and screenshots, if applicable, as well as details about compensation (if any)? 
    \item[] Answer: \answerNA{} % Replace by \answerYes{}, \answerNo{}, or \answerNA{}.
    \item[] Justification: 

\item {\bf Institutional review board (IRB) approvals or equivalent for research with human subjects}
    \item[] Question: Does the paper describe potential risks incurred by study participants, whether such risks were disclosed to the subjects, and whether Institutional Review Board (IRB) approvals (or an equivalent approval/review based on the requirements of your country or institution) were obtained?
    \item[] Answer: \answerNA{} % Replace by \answerYes{}, \answerNo{}, or \answerNA{}.
    \item[] Justification:

\item {\bf Declaration of LLM usage}
    \item[] Question: Does the paper describe the usage of LLMs if it is an important, original, or non-standard component of the core methods in this research? Note that if the LLM is used only for writing, editing, or formatting purposes and does \emph{not} impact the core methodology, scientific rigor, or originality of the research, declaration is not required.
    %this research? 
    \item[] Answer: \answerYes{} % Replace by \answerYes{}, \answerNo{}, or \answerNA{}.
    \item[] Justification: LLMs were used for dataset generation, specifically for \textsc{Shor} and \texttt{SHOR-flaw}. 
    See Appendix~\ref{app:SHOR} for details.

\end{enumerate}

\newpage
\end{document}